\documentclass[a4paper,conference]{IEEEtran}
\ifCLASSINFOpdf
\else
\fi
\usepackage{hyperref}
\usepackage[pdftex]{graphicx}
\usepackage[symbol]{footmisc}

\makeatletter
\newcommand{\printfnsymbol}[1]{%
  \textsuperscript{\@fnsymbol{#1}}%
}
\makeatother

\begin{document}
%
\title{DocEnTr: An End-to-End Document Image Enhancement Transformer}

\author{\IEEEauthorblockN{Mohamed Ali Souibgui\textsuperscript{\textsection}} 
\IEEEauthorblockA{Computer Vision Center\\
Universitat Aut\`onoma de Barcelona\\
Barcelona, Spain\\
msouibgui@cvc.uab.es} 
\and
\IEEEauthorblockN{Sanket Biswas\textsuperscript{\textsection}}
\IEEEauthorblockA{Computer Vision Center\\
Universitat Aut\`onoma de Barcelona\\
Barcelona, Spain\\
sbiswas@cvc.uab.es}  
\and
\IEEEauthorblockN{Sana Khamekhem Jemni\textsuperscript{\textsection}}
\IEEEauthorblockA{Digital Research Center of Sfax\\
MIRACL Laboratory, University of Sfax\\
Sfax, Tunisia\\
sana.khamekhem@gmail.com}
\and
\IEEEauthorblockN{Yousri Kessentini}
\IEEEauthorblockA{Digital Research Center of Sfax\\
SM@RTS Laboratory\\
Sfax, Tunisia\\
yousri.kessentini@crns.rnrt.tn}
\and
\IEEEauthorblockN{Alicia Forn\'{e}s, Josep Llad\'{o}s}
\IEEEauthorblockA{Computer Vision Center, Computer Science Dept.\\
Universitat Aut\`onoma de Barcelona\\
Barcelona, Spain\\
\{afornes, josep\}@cvc.uab.es}
\and
\IEEEauthorblockN{Umapada Pal}
\IEEEauthorblockA{CVPR Unit\\
Indian Statistical Institute\\
Kolkata, India\\
umapada@isical.ac.in}}

\maketitle

\begingroup\renewcommand\thefootnote{\textsection}
\footnotetext{Equal contribution}
\endgroup

\begin{abstract}
Document images can be affected by many degradation scenarios, which cause recognition and processing difficulties. In this age of digitization, it is important to denoise them for proper usage. To address this challenge, we present a new encoder-decoder architecture based on vision transformers to enhance both machine-printed and handwritten document images, in an end-to-end fashion. The encoder operates directly on the pixel patches with their positional information without the use of any convolutional layers, while the decoder reconstructs a clean  image from the encoded patches. Conducted experiments show a superiority of the proposed model compared to the state-of-the-art methods on several DIBCO benchmarks. Code and models will be publicly
available at: \url{https://github.com/dali92002/DocEnTR}.

\end{abstract}
\IEEEpeerreviewmaketitle

\section{Introduction}\label{s:intro}


The preservation and legibility of document images (especially the historical ones) are of utmost priority for the Document Image Analysis and Recognition (DIAR) research. Document records usually contain significant information and in the historical cases it dates back centuries and decades \cite{megyesi2019decode}.  The conservation of document records can be hampered by several kinds of degradation such as smears, stains, artefacts, pen strokes, bleed-through effects and uneven illumination. These distortions could heavily impact the subsequent downstream tasks for information processing, such as segmentation,  Optical Character Recognition (OCR), information spotting  and  layout analysis. This manifests the need for a robust pre-processing task that denoises and reconstructs a high-quality clean image from its already degraded counterpart. Document Image Enhancement (DIE) aims towards restoring the quality of the degraded document samples to yield a clear enhanced version that is locally uniform.


In recent times, Convolutional Neural Network (CNN)-based approaches have been widely applied to DIE related sub-tasks, like binarization \cite{kang2021complex,jemni2022enhance}, deblurring \cite{hradivs2015convolutional}, shadow \cite{wang2019an} and watermark removal \cite{souibgui2020gan}, etc.  Although the performance of these models has significantly improved over classical handcrafted techniques, they do have their own set of drawbacks. Firstly, CNNs operate on regular grids and using the same convolutional filter to restore different regions of a degraded document image may not be a sensible choice. Secondly, CNNs fail to capture high-level long-range dependencies as they are more suited for extracting low-level spatial information from images. 

With the recent  success of transformers in Natural Language Processing (NLP) \cite{vaswani2017attention, devlin2018bert}, its application to computer vision problems (like image recognition \cite{dosovitskiy2021an}, object detection \cite{carion2020end}, visual question answering \cite{biten2021latr}, handwritten text recognition (HTR) \cite{rouhou2021transformer}, etc.) also started getting more prominence. The self-attention mechanism proposed in \cite{vaswani2017attention} helps to capture global interactions between contextual features. Using local information combined with the knowledge of long-range global spatial arrangement is beneficial for an efficient image restoration model. This local information is often encoded in the patch content of an image and the large scale organization is contained in the redundancy of this information across the patches of the image \cite{de2019patch}. Contrary to CNNs, which process pixel arrays, Vision Transformers (ViTs) \cite{dosovitskiy2021an} split an image into fixed-size patches (eg. 8x8, 16x16 etc.), they correctly embeds each of them as latent representation, and include positional embedding information as input to the transformer encoder. This allows to encode the relative location of the patches, along with both local (spatial) and global (semantic) long-range dependencies. The motivation of using ViTs for our overall proposed baseline model is that a missing/degraded patch in the distorted document image can be recovered from the neighbouring patches information with the power of the multi-head self-attention in ViTs, which quantifies pairwise global reasoning between them. Also, ViTs have been adapted in the overall model pipeline in an encoder-decoder based setting, inspired by the concept of denoising autoencoders \cite{vincent2008extracting} used in reconstruction of corrupted input data. The encoder is mapping the degraded image patches into latent representations, whereas the decoder is recovering a clean image version from those encoded representations.

The overall contributions of our work can be summarized into three folds:
\begin{itemize}
    \item 
    We introduce a simple and flexible Document image Enhancement Transformer (DocEnTr), an end-to-end image enhancement approach, that effectively restores and enhances a degraded document image provided as input. As far as we know, DocEnTr is the first pure transformer-based baseline that leverages the effectiveness of Vision Transformers (ViTs) in an encoder-decoder based framework, without any dependency on CNNs.   
    \item 
    We have addressed document binarization as the key problem study in this work to investigate the power of DocEnTr architecture. Experimental evaluation shows that DocEnTr achieves state-of-the-art results on standard document binarization benchmarks (DIBCO), for both machine-printed and handwritten degraded document images.   
    \item 
    A comprehensive and intuitive case study has been dedicated in Section \ref{s:results} to prove the utility of ViTs with its multi-headed self-attention mechanism in the task of document enhancement.   
\end{itemize}

The rest of this paper is organized as follows. In Section \ref{s:sota} we review the state of the art. The Document image Enhancement Transformer (DocEnTr) is described in Section \ref{s:method}. Section \ref{s:results} contains an analysis of the extensive experimentation that has been conducted, including different quantitative and qualitative studies. Finally, in Section \ref{s:conclusion} we draw the conclusions and propose open challenges for future research directions.



\section{Related Work}\label{s:sota}

\subsection{Document Image Enhancement}

This work is an application within the DIE, which has been an active filed within the DIAR community. The first classic methods were based on thresholding, which means finding a single (global) or multiple (local) threshold(s) value(s) for the  document.  These threshold values are  used to classify the document image pixels into foreground (black) or background (white) \cite{otsu1979threshold,sauvola2000adaptive}. These methods are still evolving in the recent years using machine learning tools, for instance, with support vector machines (SVM)  \cite{xiong2018degraded}. Later, energy based methods were introduced. These are based on tracking the text pixels by maximizing its energy function \cite{hedjam2014constrained}, while minimizing the one of the  degraded background. However, the results using those   approaches were unsatisfactory \cite{Pratikakis2017}.

Recently, deep learning based methods were used to tackle this problem by learning the enhancement  directly from raw data.  In \cite{afzal2015document},  the problem was formulated  as pixels classification. Each pixel is classified as black or white depending on a sequence of the surrounding pixels, where  a  2D Long Short-Term Memory (LSTM) was trained for this task.   This process is, of course, time consuming.  A more practical solution is  to map the images from the degraded domain  to the enhanced one in an end-to-end  fashion with CNN auto-encoders. These latter, hence, were leading  the recent improvements in image denoising \cite{mao2016image} and more particularly documents enhancement tasks, like binarization \cite{lore2017llnet,calvo2019selectional,akbari2020binarization},  deblurring problems \cite{hradivs2015convolutional} and so on. Following this strategy,   a fully CNN model was proposed in \cite{tensmeyer2017document} to binarize the degraded document images at multiple image scales. Similarly,  \cite{kang2021complex} proposed an auto-encoder architecture that performs a cascade of pre-trained U-Net models \cite{ronneberger2015u} to learn the binarization  using less amount of data. Moreover, generation models (GAN) were employed for this task to generate clean images by conditioning on  the degraded versions.  These architectures are composed of  a generative model that produces a clean version of the image and a discriminator to assess the binarization result. Both models are usually composed of fully (or partially) CNN layers.   In \cite{souibgui2020gan}, a conditional GAN approach was proposed for different  enhancement tasks achieving good results in  document images cleaning, binarization,  deblurring and dense watermarks removal. This method was recently extended in \cite{jemni2022enhance} by adding  a second discriminator  to assess the text readability  for the goal of  obtaining an enhanced  image that is  clean and readable at the same time.   A similar cGAN's based method was also proposed in \cite{zhao2019document, bhunia2019improving, tamrin2021, souibgui2020conditional}. 

\subsection{Transformers in Vision and Image Enhancement Tasks}

In the very recent years, transformers are behind the  advances in deep learning applications. Transformer based architectures firstly showed a great success in NLP tasks \cite{vaswani2017attention, devlin2018bert} for text translation and embedding, surpassing the previous LSTM approaches. This motivates many works  to employ them  for the vision tasks, for instance, classification \cite{dosovitskiy2021an}, object detection \cite{carion2020end},  document understanding \cite{xu2020layoutlmv2, appalaraju2021docformer, li2021selfdoc}, etc.  More related to this paper, transformers were also used for natural image restoration \cite{liang2021swinir} and document images dewarping \cite{feng2021doctr}. However, the architectures that were used in these later image and document enhancement  approaches are still relying on the CNN feature extractors before passing to the transformers stage. Also, the CNN are used to reconstruct the output image.  Contrary,  what we are proposing in this work is a fully transformer approach that attends directly to the patches on the input images and reconstruct the pixels without the using of any CNN layer.

\section{Method}\label{s:method}
\begin{figure*}[!t]
\centering
\includegraphics[width=\linewidth]{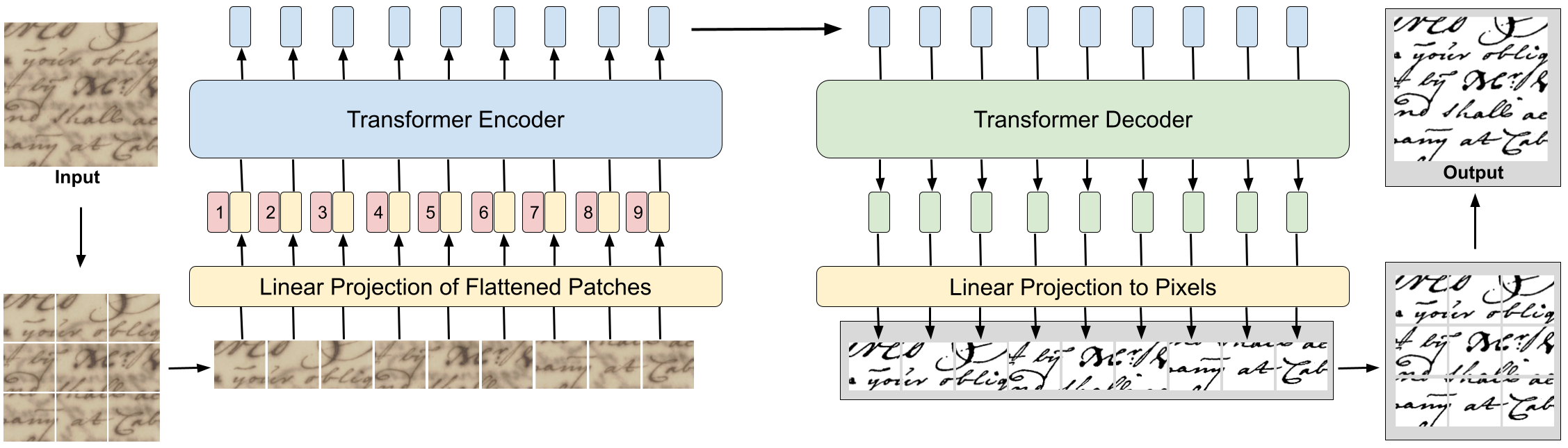}
\caption{Proposed model: The input image is split into patches, which are linearly embedded, and the position information are added to them. The resulting sequence of vectors are fed to a standard Transformer encoder to obtain the latent representations. These representations are fed to another Transformer representing the decoder to obtain the decoded vector, which is linearly projected to vectors of pixels representing the output image patches.}
\label{fig:architecture}
\end{figure*}

The proposed model is a scalable auto-encoder that uses vision transformers in its encoder and decoder parts, as illustrated in Fig~\ref{fig:architecture}.  The degraded image is first divided into   patches before entering to the encoder part. During encoding, the patches are mapped  to a latent representation of tokens,  where each token is associated with a degraded patch.  Then, the tokens are  passed to the  decoder that outputs  the  enhanced version of patches. Unlike the  CNN based auto-encoders,  which were usually employed for the document image enhancement tasks, the transformer auto-encoder is profiting from the self attention mechanism which gives a global information during every patch enhancement. Both  decoder and especially encoder are inspired from the vision transformer (ViT) \cite{dosovitskiy2021an} architecture.  We present more details of the model's architecture  in what follows.

\subsection{Encoder}
In the encoding stage (left part of Fig.\ref{fig:architecture}),  given an image, we divide it into a set of patches. Then, we embed these  patches to obtain the tokens and add their positional information. After that, a number of transformer blocks  is employed to map these tokens into the encoded latent representation.  These blocks follow the same structure as \cite{dosovitskiy2021an}, composed of alternating layers of multi-headed self-attention and multi-layered perceptron (MLP). Each of these blocks are preceded by a LayerNorm (LN) \cite{ba2016layer}, and followed by a residual connection. The patches embedding size and the number of transformer blocks are set depending on the model size.

\subsection{Decoder}

The decoder part consists of a series of transformer blocks (having the same number as the encoder blocks) that take as an input the sequence of  outputted tokens from the encoder. These tokens are propagated in the transformer decoder blocks, and then  projected with a linear layer to the desired pixel values.  This makes each element of the output correspond to a vector representing a flattened patch in the output image. The ground truth pixel values are obtained by dividing the ground truth (GT) clean image into patches (in the same way as the input degraded image) and flattening them into vectors. A mean squared error (MSE) loss is used between the model's output and the GT pixel patches to train the model.

\subsection{Model Variants}

Following a similar convention as previous works \cite{devlin2018bert,dosovitskiy2021an}, the proposed model configuration can be modified to produce different variants. In our experiments we define three types of variants which are "Small", "Base" and "Large", as enlisted in Table~\ref{table:models_variants}. Evidently, setting a larger model require more computational memory and training time since the number of model parameters is increasing. Thus, a trade off between the model size and its enhancement performance must be taken into consideration.

\begin{table}[h]
\renewcommand{\arraystretch}{1.3}
\caption{Details of our model variants}
\label{table:models_variants}
\centering
\begin{tabular}{|c|c|c|c|c|}
\hline
Model& Layers& Dim& Attention Heads & \# Parameters  \\
\hline
DocEnTr-Small & 6 & 512 & 4 & 17M\\
\hline
DocEnTr-Base & 12 & 768 & 8 & 68M\\
\hline
DocEnTr-Large & 24 & 1024 & 16 & 255M\\
\hline
\end{tabular}
\end{table}

\section{Experimental Validation}\label{s:results}
To validate our model, we use the datasets proposed in the different DIBCO and H-DIBCO contests \cite{Pratikakis2018icfhr} for printed and handwritten degraded document images binarization and compare our results with the state of the art methods. Before these experiments, we have conducted different investigations for a proper selection of the hyperparameters.  

\subsection{Choosing the Best Model Configuration}

We begin our experiments by choosing the configuration that gives the best performance from our model variants (Small, Base or Large). For training, each degraded image and its GT clean one is divided into overlapped patches with sizes $256\times256\times3$, the overlapping was set vertically and horizontally by a half of the patches size (means 128). These resultant images (patches) will be used by our models as an input and expected output (training data).  For results evaluation, and same as the usual approaches \cite{Pratikakis2010ICFHR}, we utilize the following metrics: Peak signal-to-noise ratio (PSNR), F-Measure (FM), pseudo-F-measure (F$_{ps}$) and Distance reciprocal distortion metric (DRD). 
We used in this experiment the DIBCO 2017 dataset, and the obtained results are  given in Table~\ref{table:models_size_test}. As it can be seen, a larger model gives a better result in all the metrics, but it requires more computation resources. Thus, we recommend using a Base model for a binarization task. Nevertheless, we will test as well the Large version in following experiments. 

\begin{table}[h]
\renewcommand{\arraystretch}{1.3}
\caption{Results of varying the model size for the DIBCO 2017 dataset. $\uparrow$: The higher the better. $\downarrow$: The lower the better.}
\label{table:models_size_test}
\centering
\begin{tabular}{|c|c|c|c|c|}
\hline
Model& PSNR $\uparrow$ &FM $\uparrow$& F$_{ps}$ $\uparrow$& DRD $\downarrow$ \\
\hline
DocEnTr-Small & 18.29 & 91.06 & 93.82 & 2.78\\
\hline
DocEnTr-Base & 18.69 & 91.66 & 94.11 & 2.63 \\
\hline
DocEnTr-Large & \textbf{18.85} & \textbf{92.14} & \textbf{94.58} & \textbf{2.53}\\
\hline
\end{tabular}
\end{table}

Next, we do another experiment related to the input image size, and the patches size that are used by our model. The reason behind is that having different image size and patch size can affect the binarization since the model is accessing to different type of information (from global to local). The obtained results using the Base model are given in Table~\ref{table:input_size_test}. As it can be seen, a slightly better performance is obtained using an input with the smaller size ($256\times256\times3$ compared to $512\times512\times3$ ). However, we can notice that the performance is highly improved when using a smaller patch size. The reason is that, by employing a smaller patch size, we make each patch of the image attending to more and much local patches during the self-attention. Thus, the model is looking to more and much fine  information during the enhancement process with $8\times8$ patch size. But, as before, using a smaller patch size means augmenting the model parameters, requiring more computation resources.

\begin{table}[h]
\renewcommand{\arraystretch}{1.3}
\caption{Results of varying the input and patch sizes for the DIBCO 2017 dataset}
\label{table:input_size_test}
\centering
\begin{tabular}{|c|c|c|c|c|c|}
\hline
Input Size& Patch Size & PSNR $\uparrow$ &FM $\uparrow$& F$_{ps}$ $\uparrow$& DRD $\downarrow$ \\
\hline
$256\times256\times3$ &$8\times8$ & \textbf{19.11} & \textbf{92.53} & \textbf{95.15} & \textbf{2.37}\\
\hline
$256\times256\times3$ &$16\times16$ & 18.69& 91.66&	94.11& 2.63\\
\hline
$256\times256\times3$ &$32\times32$ & 17.57&	89.37&	91.99&	3.44\\
\hline
$512\times512\times3$ &$8\times8$ & 18.91 &	92.2&	94.93&	2.45\\
\hline
$512\times512\times3$ &$16\times16$ & 18.66&	92.15&	93.89&	2.54\\
\hline
$512\times512\times3$ &$32\times32$ & 17.27&	89.43&	91.51&	3.54\\
\hline
\end{tabular}
\end{table}

\subsection{Quantitative Evaluation}

After choosing the best hyper-parameters of the model, we conduct the experiments on the different datasets and compare our result with the related approaches. We begin by testing with the DIBCO 2011 dataset  \cite{Pratikakis2011icdar}. This dataset contains degraded document images  with handwritten and printed text. For training, we use all the images from the other DIBCO and H-DIBCO datasets (except DIBCO 2019) and the Palm Leaf dataset \cite{burie2016icfhr2016}. These images are split into overlapped images with size $256\times256\times3$ before being fed to the model. The obtained results are given in Table~\ref{table:dibco2011}, where we can notice a superiority of out method compared to the different variations of the related approaches. We choose to compare with different families of approaches: classic thresholding and deep learning based methods (whether basing on CNN or cGAN). Our model DocEnTr-Base\{8\}, which means using the Base setting with a patch size of $8\times8$, gives the best PSNR and DRD compared to all the other methods. While the model DocEnTr-Large\{16\}, which means using the Large setting with a patch size of $16\times16$, leads to the second best performance in the metrics PSNR, F$_{ps}$ and DRD. We note that for a computation reason, we were not able to train the Large setting with a patch size of  $8\times8$.

\begin{table}[h]
\renewcommand{\arraystretch}{1}
\caption{Comparative results of our proposed method on DIBCO 2011 Dataset. Thresh: Thresholding, Tr: Transformers.}
\label{table:dibco2011}
\centering
\begin{tabular}{|c|c|c|c|c|c|}
\hline
Method& Model &PSNR $\uparrow$ &FM $\uparrow$& F$_{ps}$ $\uparrow$& DRD $\downarrow$ \\
\hline
Otsu \cite{otsu1979threshold}& Thres. & 15.70 & 82.10 & -- & 9.00\\
\hline
Savoula et al. \cite{sauvola2000adaptive}& Thres. & 15.60& 82.10 & -- & 8.50\\
\hline
Vo et al. \cite{Vo2018} & CNN & 20.10& 93.30 & -- & 2.00\\
\hline
Kang et al \cite{kang2021complex} & CNN & 19.90& \textbf{95.50} & -- & 1.80\\
\hline
Tensmeyer et al \cite{tensmeyer2017document} & CNN & 20.11& 93.60 & \textbf{97.70} & 1.85\\
\hline
Zhao et al. \cite{Vo2018} &cGAN & 20.30& 93.80 & -- & 1.80\\
\hline
\textbf{DocEnTr-Base\{8\} }& Tr& \textbf{20.81}& 94.37& 96.15& \textbf{1.63}\\
\hline
\textbf{DocEnTr-Base\{16\}}& Tr  & 20.11&	93.48&	96.12&	1.93\\
\hline
\textbf{DocEnTr-Large\{16\}}& Tr & 20.62& {94.24}& 96.71& 1.69 \\
\hline

\end{tabular}
\end{table}

After that, we test our model on the H-DIBCO 2012 dataset \cite{Pratikakis2012ICFHR}, which contains degraded handwritten document images. As in the previous experiment, we use the other datasets for training with the same split size.   The obtained results are shown in Table~\ref{table:dibco2012}, where we can notice that our model gives the best performance in terms of PSNR and FM with the Base\{8\} configuration. We notice also that the other configuration gives competitive results compared to the other approaches.

\begin{table}[h]
\renewcommand{\arraystretch}{1}
\caption{Comparative results of our proposed method on H-DIBCO 2012 Dataset. Thresh: Thresholding, Tr: Transformers.}
\label{table:dibco2012}
\centering
\begin{tabular}{|c|c|c|c|c|c|}
\hline
Method& Model &PSNR $\uparrow$ &FM $\uparrow$& F$_{ps}$ $\uparrow$& DRD $\downarrow$ \\
\hline
Otsu \cite{otsu1979threshold}& Thres. & 15.03 & 80.18 &82.65 & 26.46\\
\hline
Savoula et al. \cite{sauvola2000adaptive}& Thres.& 16.71 & 82.89& 87.95& 6.59\\
\hline
Kang et al \cite{kang2021complex} & CNN & 21.37& 95.16& 96.44& \textbf{1.13}\\
\hline
Tensmeyer et al \cite{tensmeyer2017document} & CNN & 20.60& 92.53 & \textbf{96.67} & 2.48\\
\hline
Zhao et al. \cite{Vo2018} &cGAN & 21.91 & 94.96& 96.15& 1.55\\
\hline
Jemni et al. \cite{jemni2022enhance} &cGAN & 22.00& 95.18& 94.63& 1.62\\
\hline
\textbf{DocEnTr-Base\{8\} }& Tr& \textbf{22.29}& \textbf{95.31} & 96.29& 1.60\\
\hline
\textbf{DocEnTr-Base\{16\}}& Tr  & 21.03& 93.31 & 94.72& 2.31\\
\hline
\textbf{DocEnTr-Large\{16\}}& Tr & 22.04& 95.09& 96.00 & 1.64 \\
\hline

\end{tabular}
\end{table}

Moreover, we tested with the more recent DIBCO 2017 dataset. In this dataset our model achieves the best performance in all the evaluation metrics, as presented in Table~\ref{table:dibco2017}.

\begin{table}[h]
\renewcommand{\arraystretch}{1}
\caption{Comparative results of our proposed method on DIBCO 2017 Dataset. Thresh: Thresholding, Tr: Transformers.}
\label{table:dibco2017}
\centering
\begin{tabular}{|c|c|c|c|c|c|}
\hline
Method& Model &PSNR $\uparrow$ &FM $\uparrow$& F$_{ps}$ $\uparrow$& DRD $\downarrow$ \\
\hline
Otsu \cite{otsu1979threshold}& Thres.&13.85 & 77.73 &77.89 & 15.54\\
\hline
Savoula et al. \cite{sauvola2000adaptive}& Thres.&14.25 & 77.11& 84.1 & 8.85\\
\hline
Kang et al \cite{kang2021complex} & CNN& 15.85 & 91.57& 93.55 & 2.92\\
\hline
Competition top \cite{Pratikakis2017} & CNN & 18.28 & 91.04 & 92.86 & 3.40\\
\hline
Zhao et al. \cite{Vo2018} &cGAN & 17.83 & 90.73& 92.58 & 3.58\\
\hline
Jemni et al. \cite{jemni2022enhance} &cGAN & 17.45 & 89.8&  89.95&  4.03\\
\hline
\textbf{DocEnTr-Base\{8\} }& Tr& \textbf{19.11} & \textbf{92.53} & \textbf{95.15} & \textbf{2.37}\\
\hline
\textbf{DocEnTr-Base\{16\}}& Tr  & 18.69 & 91.66 & 94.11 & 2.63\\
\hline
\textbf{DocEnTr-Large\{16\}}& Tr & 18.85 & 92.14 & 94.58 & 2.53 \\
\hline

\end{tabular}
\end{table}

Lastly, we test on the H-DIBCO 2018 dataset. Here, as shown in Table~\ref{table:dibco2018}, the best performance is achieved by \cite{jemni2022enhance} basing on cGAN. Anyway, we can notice that our model is still very competitive since it ranks second in the PSNR, FM and F$_{ps}$ metrics.

\begin{table}[h]
\renewcommand{\arraystretch}{1}
\caption{Comparative results of our proposed method on DIBCO 2018 Dataset. Thresh: Thresholding, Tr: Transformers.}
\label{table:dibco2018}
\centering
\begin{tabular}{|c|c|c|c|c|c|}
\hline
Method& Model &PSNR $\uparrow$ &FM $\uparrow$& F$_{ps}$ $\uparrow$& DRD $\downarrow$ \\
\hline
Otsu \cite{otsu1979threshold}& Thres.& 9.74 &51.45 & 53.05  & 59.07\\
\hline
Savoula et al. \cite{sauvola2000adaptive}& Thres.&  13.78 &67.81& 74.08  &  17.69\\
\hline
Kang et al \cite{kang2021complex} & CNN & 19.39 & 89.71 & 91.62  & \textbf{2.51}\\
\hline
Competition top \cite{Pratikakis2017} & CNN & 19.11 & 88.34 & 90.24 & 4.92\\
\hline
Zhao et al. \cite{Vo2018} &cGAN & 18.37 & 87.73 &  90.60 & 4.58\\
\hline
Jemni et al. \cite{jemni2022enhance} &cGAN & \textbf{20.18} & \textbf{92.41} & \textbf{94.35} & 2.60\\
\hline
\textbf{DocEnTr-Base\{8\} }& Tr& 19.46 & 90.59 & 93.97 & 3.35\\
\hline
\textbf{DocEnTr-Base\{16\}}& Tr  & 19.33 & 89.97 & 93.5 & 3.68\\
\hline
\textbf{DocEnTr-Large\{16\}}& Tr & 19.47 & 89.21 & 92.54 & 3.96 \\
\hline

\end{tabular}
\end{table}

To summarize the quantitative evaluation, we demonstrate that our model gives good results compared to the state of the art approaches. This was shown by obtaining the best results in most of the evaluation metrics with the  H-DIBCO 2011, DIBCO 2012 and DIBCO 2017 benchmarks. 

\subsection{Qualitative Evaluation}

\begin{figure}[!t]
  \begin{center}
  
 \begin{tabular}{ccc}
 
    \includegraphics[width=0.30 \columnwidth, height=20mm]{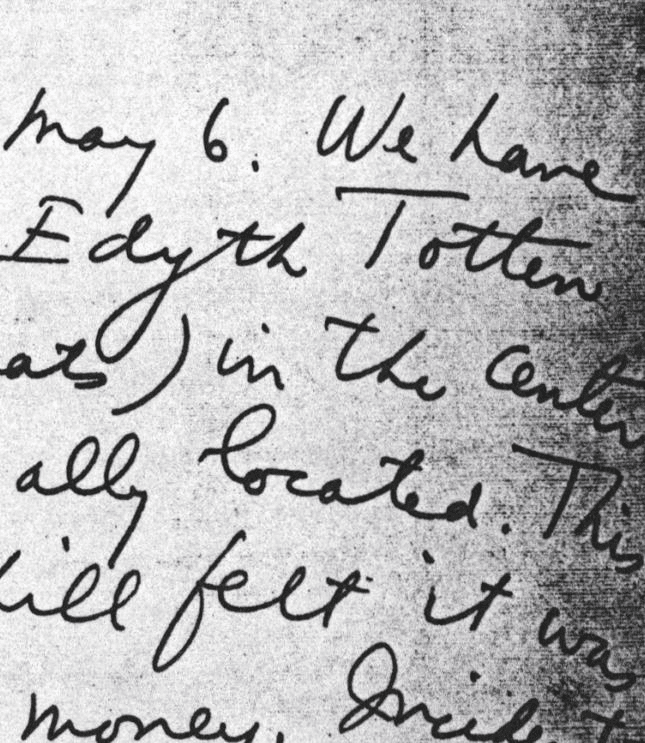} &
    \includegraphics[width=0.30 \columnwidth, height=20mm]{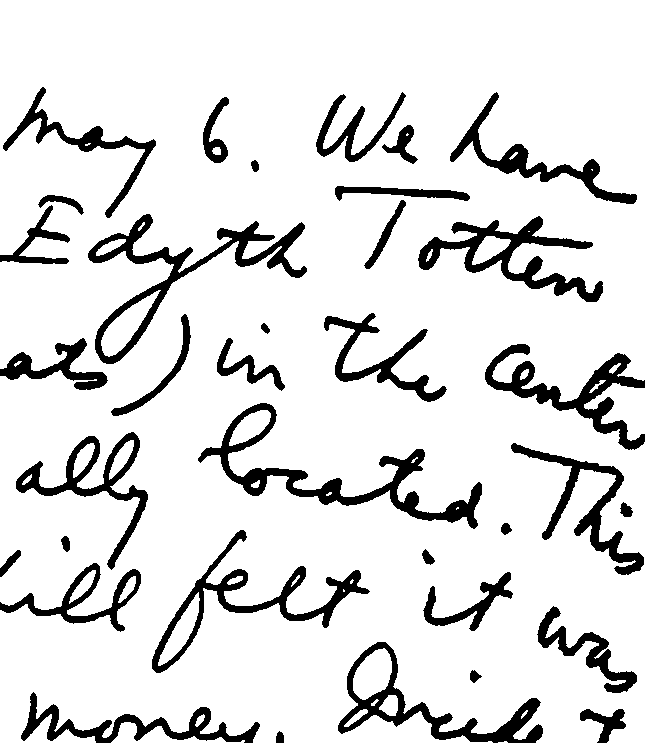}&
    \includegraphics[width=0.30 \columnwidth, height=20mm]{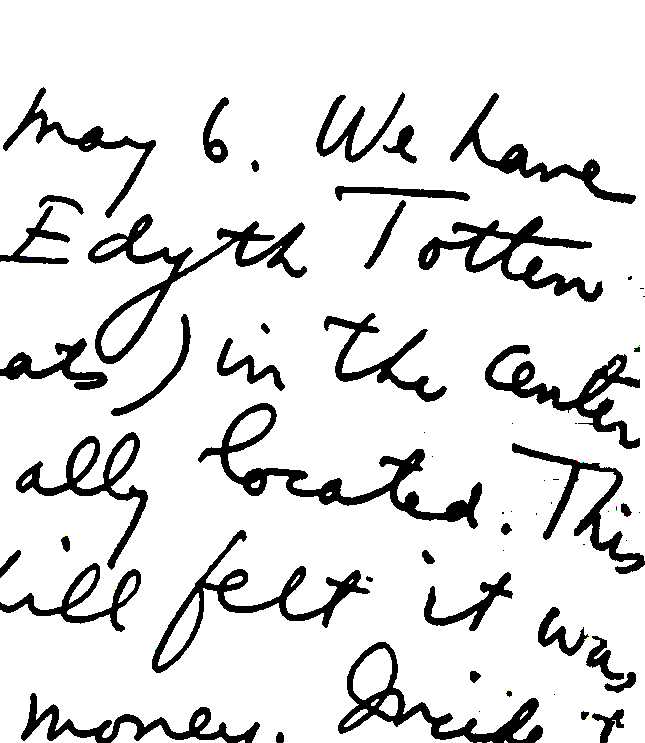}
    \\ \noalign{\smallskip} 
    
    \includegraphics[width=0.30 \columnwidth, height=20mm]{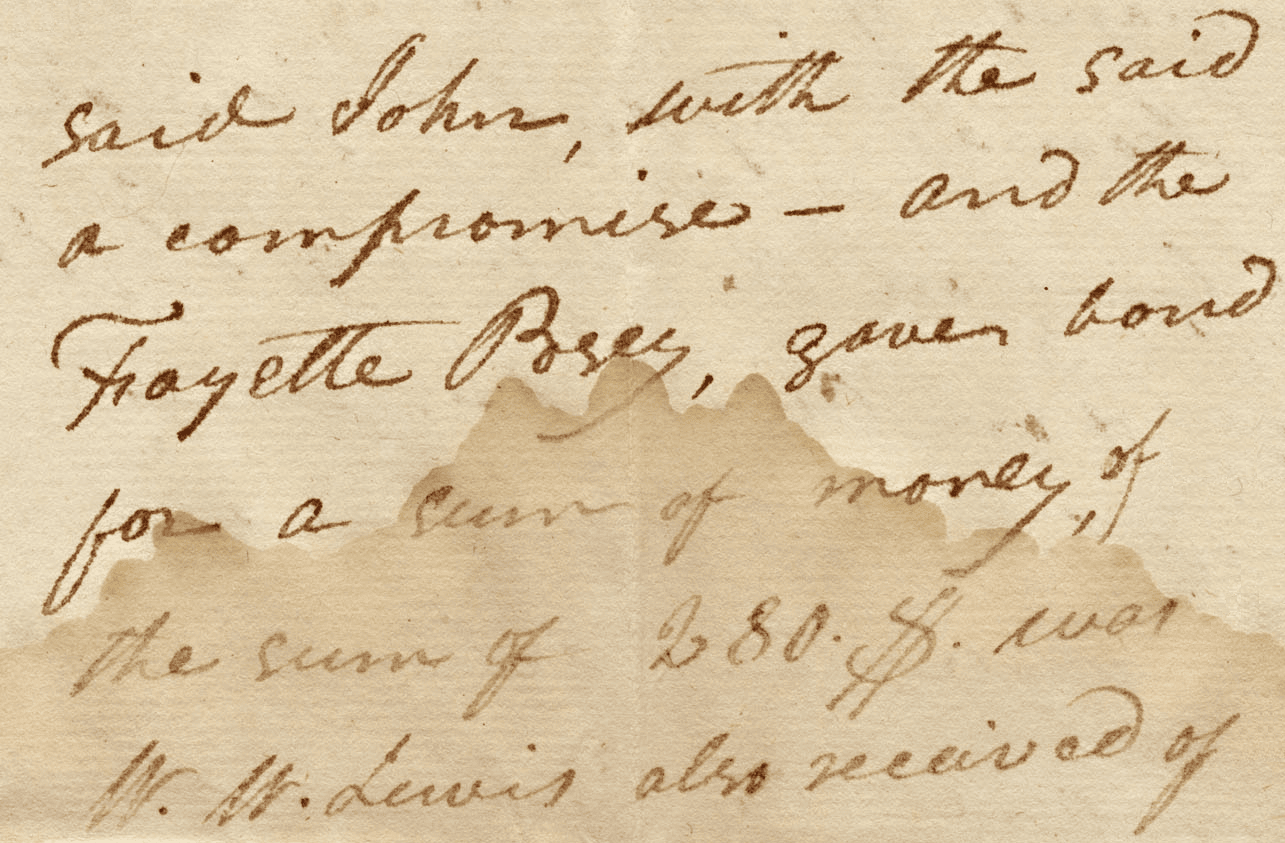} &
    \includegraphics[width=0.30 \columnwidth, height=20mm]{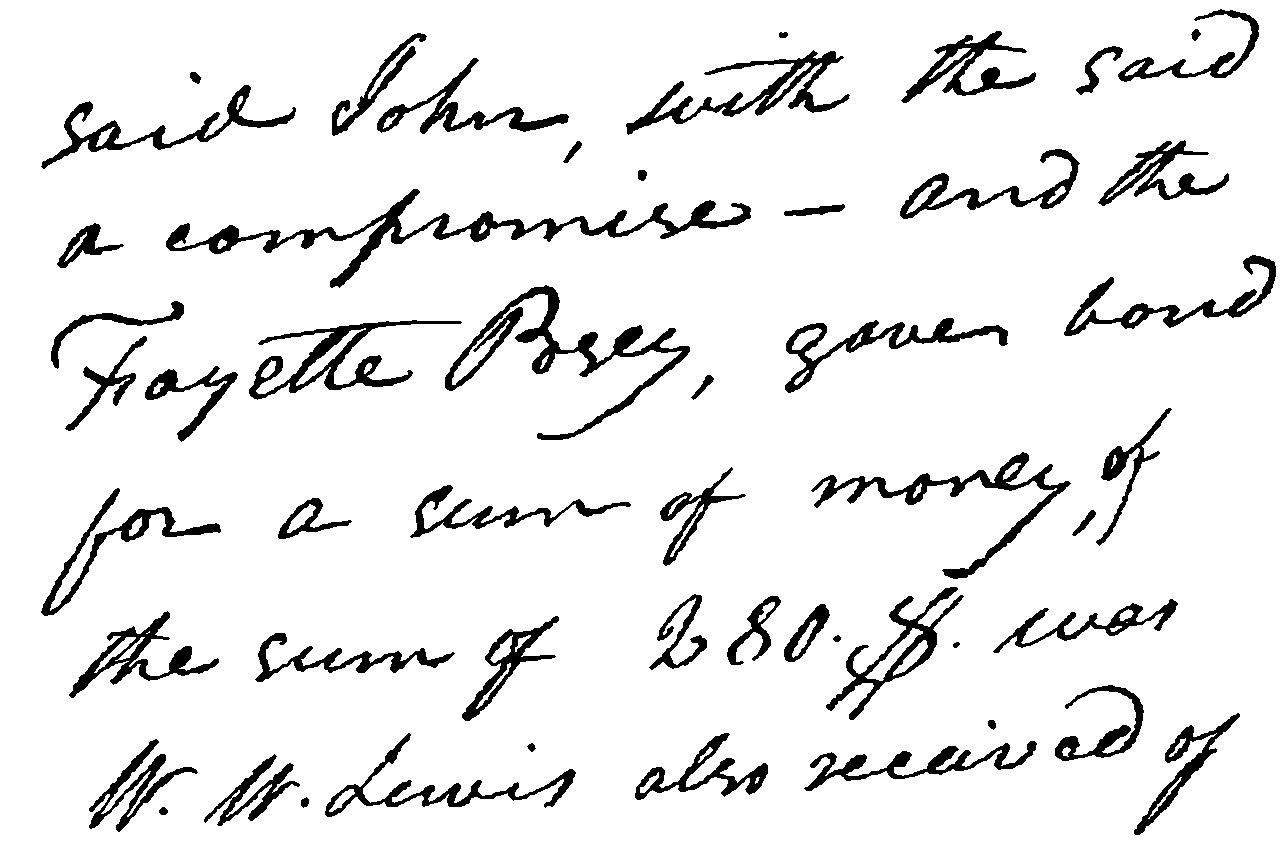}&
    \includegraphics[width=0.30 \columnwidth, height=20mm]{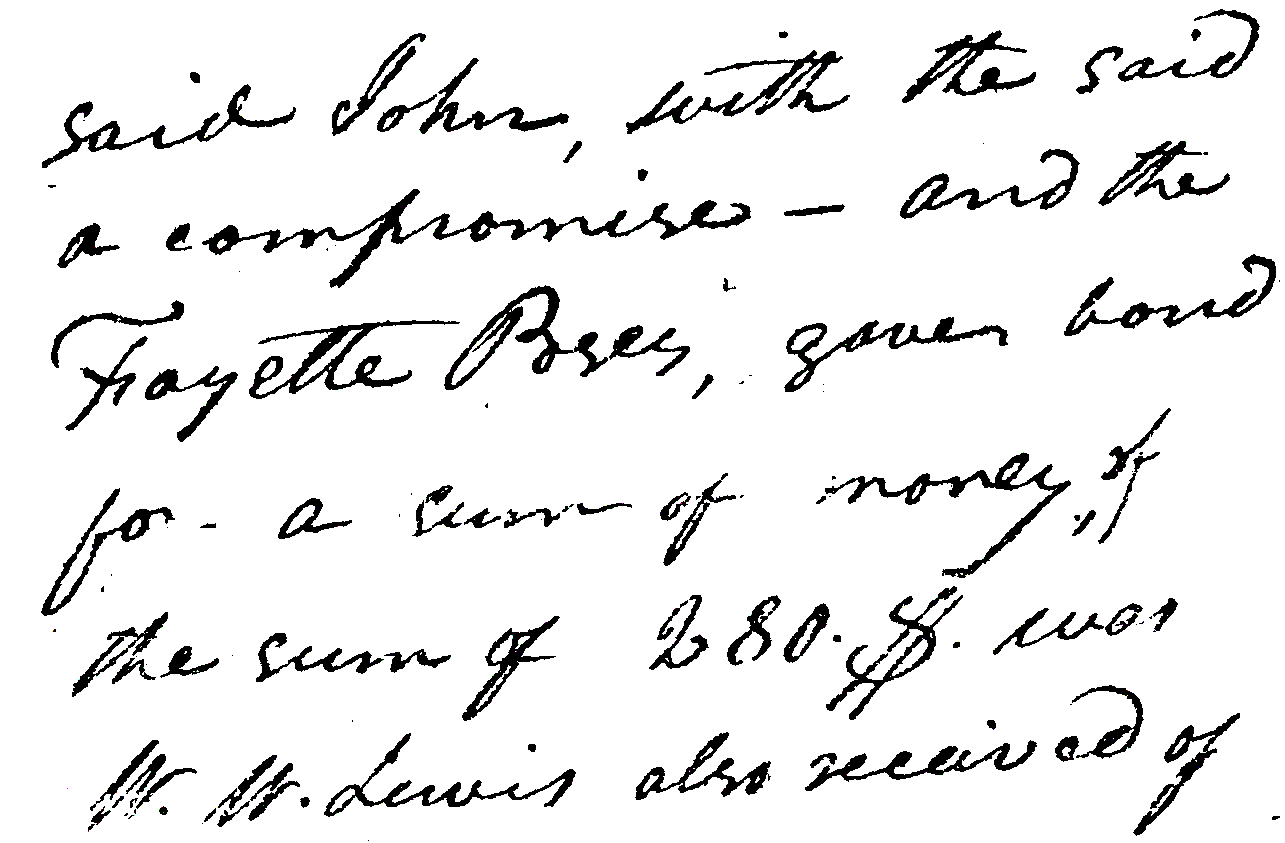}
    \\ \noalign{\smallskip}    
    
    \includegraphics[width=0.30 \columnwidth, height=20mm]{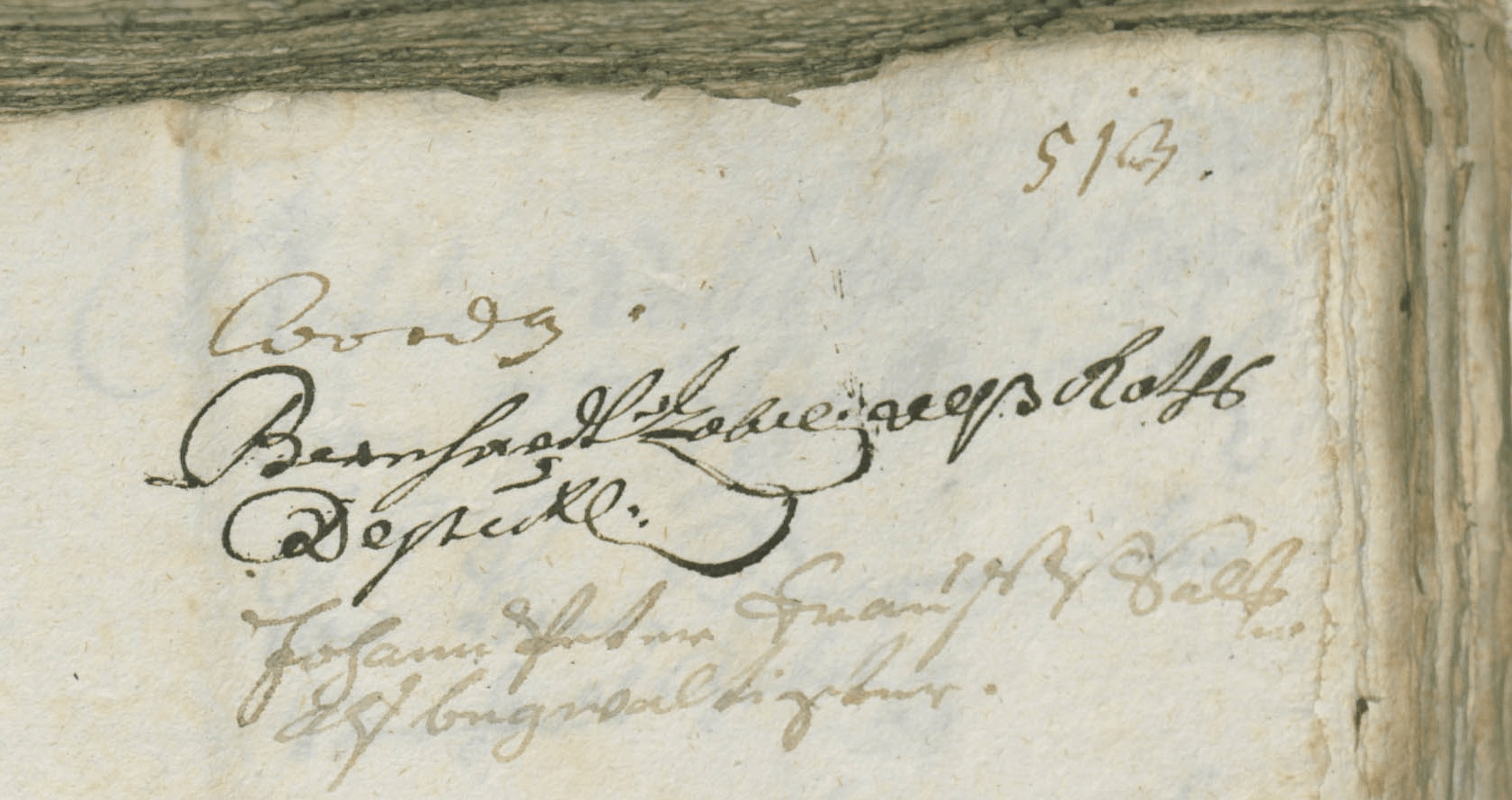} &
    \includegraphics[width=0.30 \columnwidth, height=20mm]{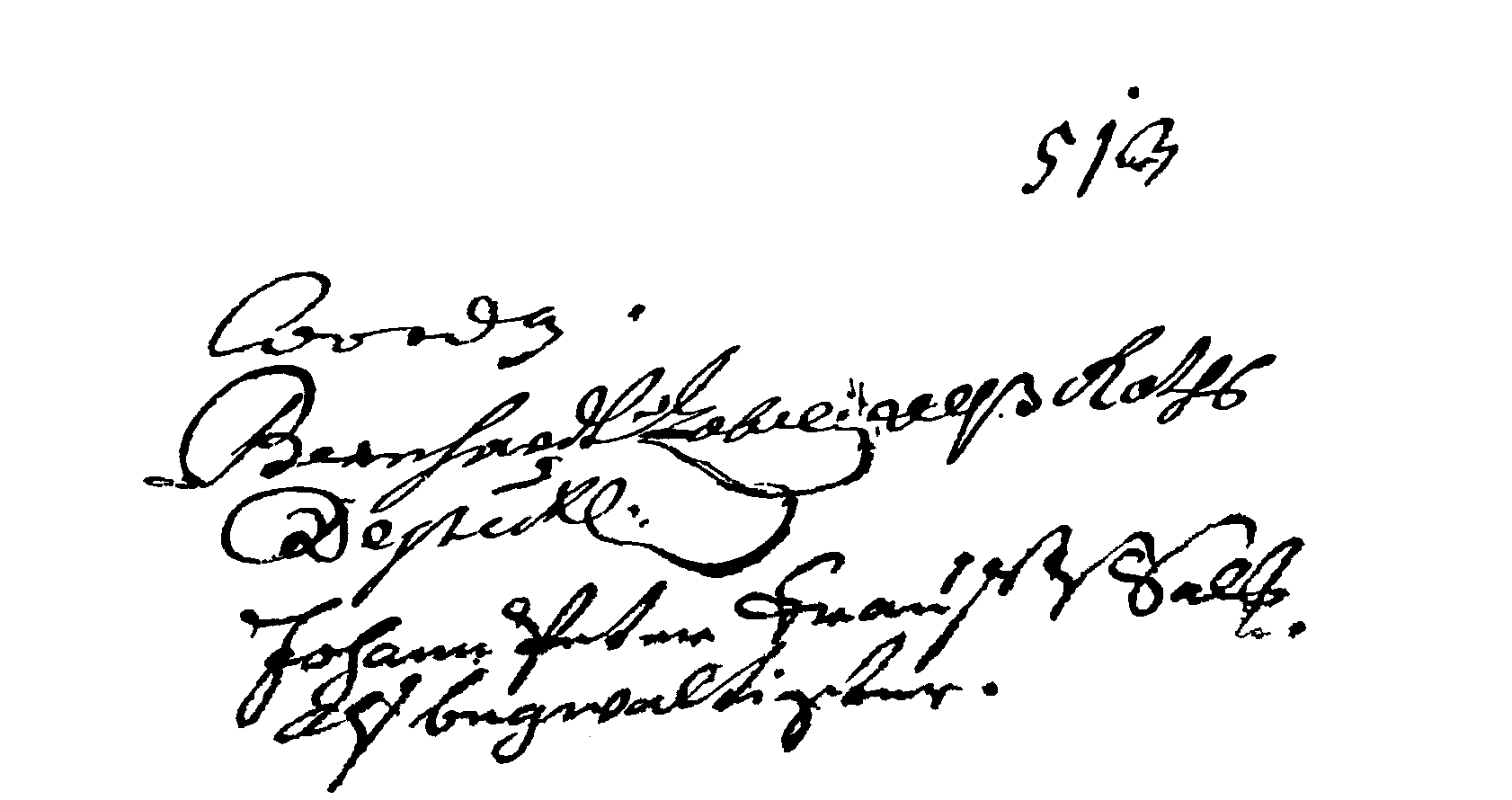}&
    \includegraphics[width=0.30 \columnwidth, height=20mm]{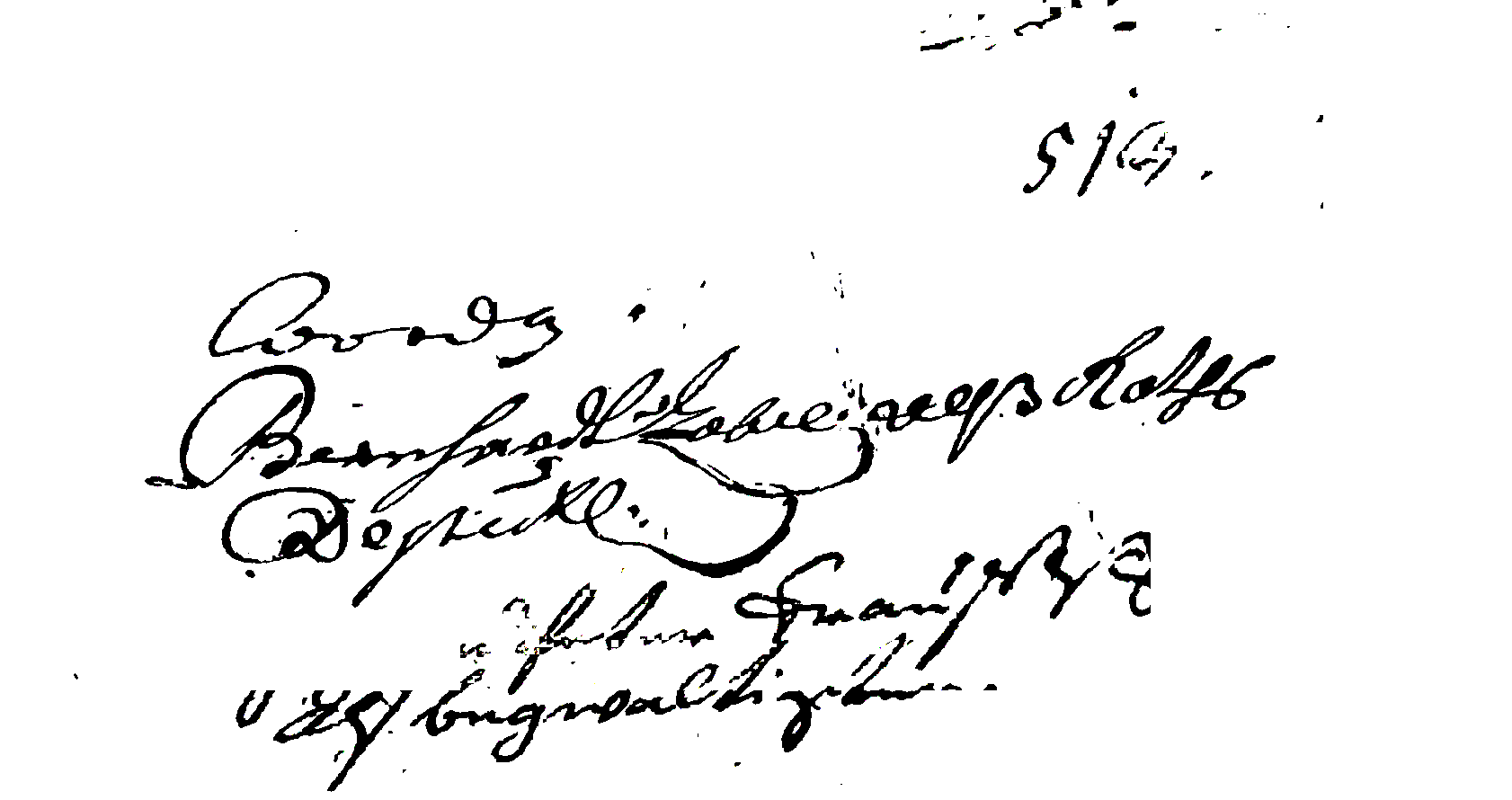}
    \\ \noalign{\smallskip}  
    
    \includegraphics[width=0.30 \columnwidth, height=20mm]{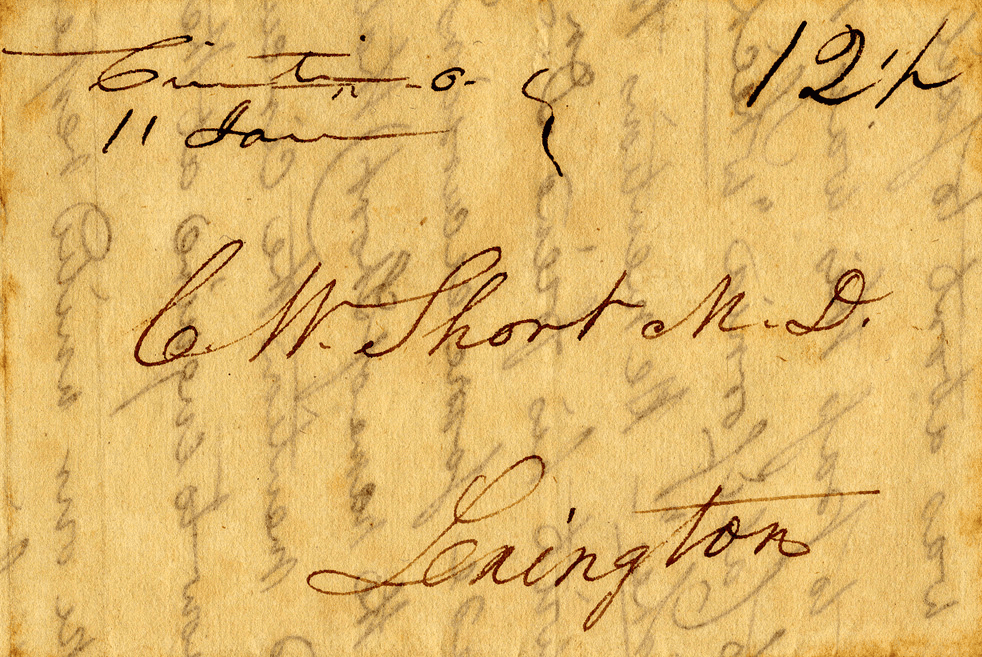} &
    \includegraphics[width=0.30 \columnwidth, height=20mm]{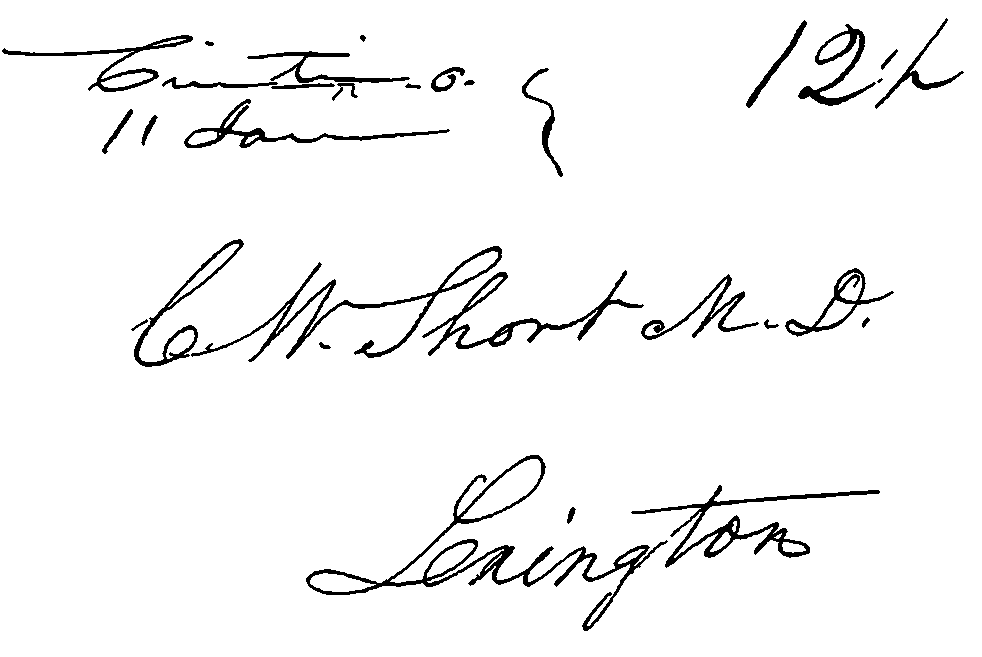}&
    \includegraphics[width=0.30 \columnwidth, height=20mm]{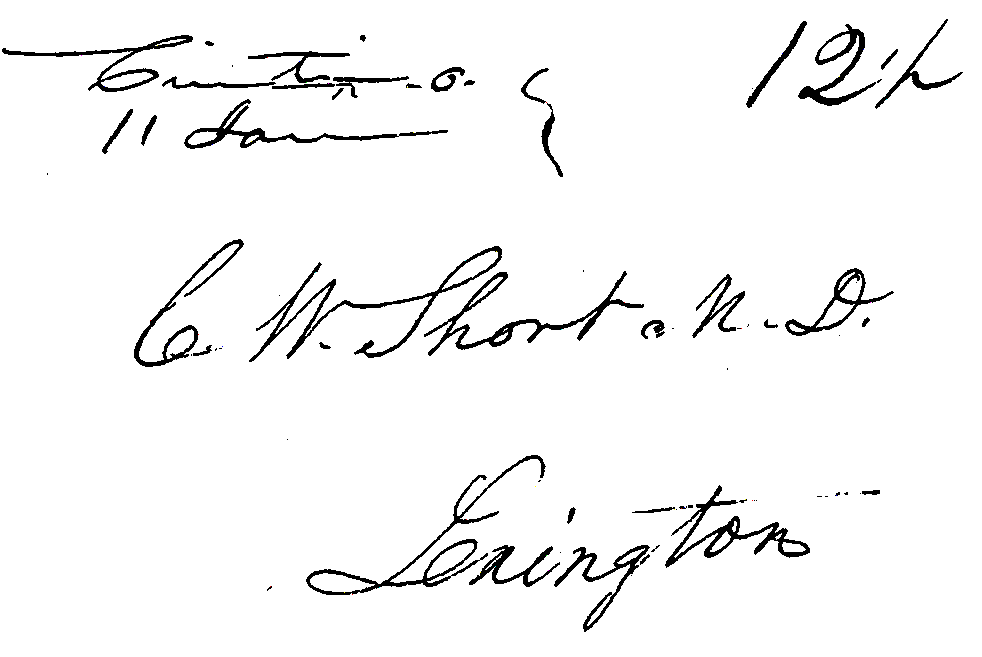}
    \\ \noalign{\smallskip}  
    
    \includegraphics[width=0.30 \columnwidth, height=20mm]{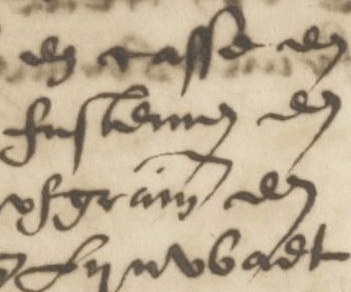} &
    \includegraphics[width=0.30 \columnwidth, height=20mm]{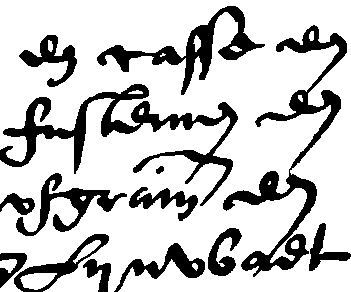}&
    \includegraphics[width=0.30 \columnwidth, height=20mm]{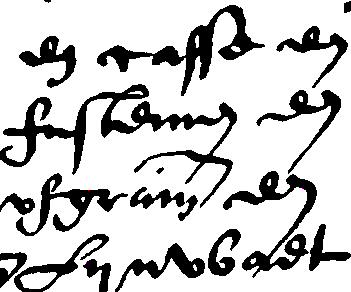}
    \\ \noalign{\smallskip}  
    
    \includegraphics[width=0.30 \columnwidth, height=20mm]{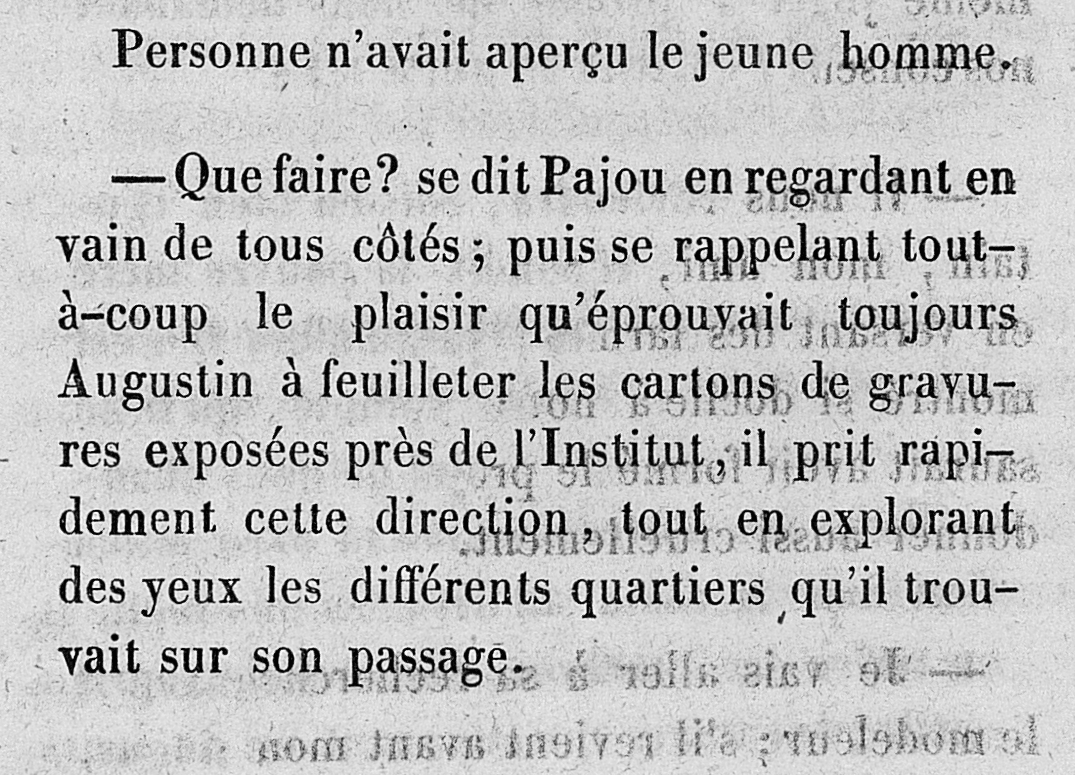} &
    \includegraphics[width=0.30 \columnwidth, height=20mm]{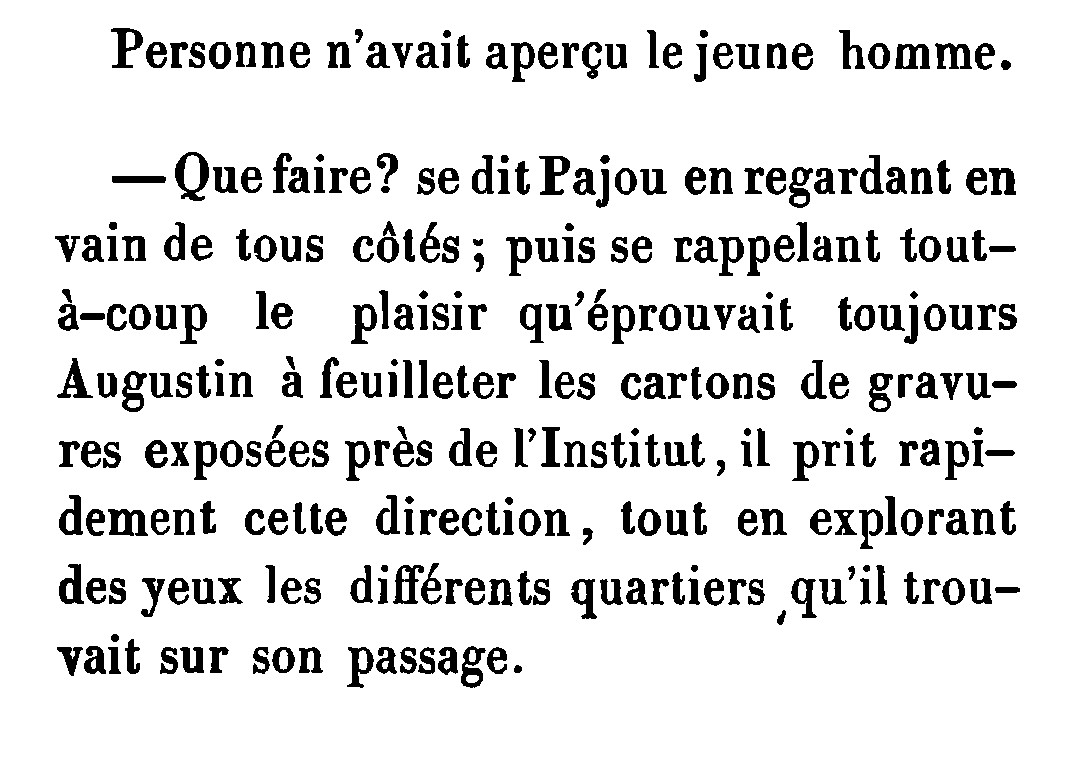}&
    \includegraphics[width=0.30 \columnwidth, height=20mm]{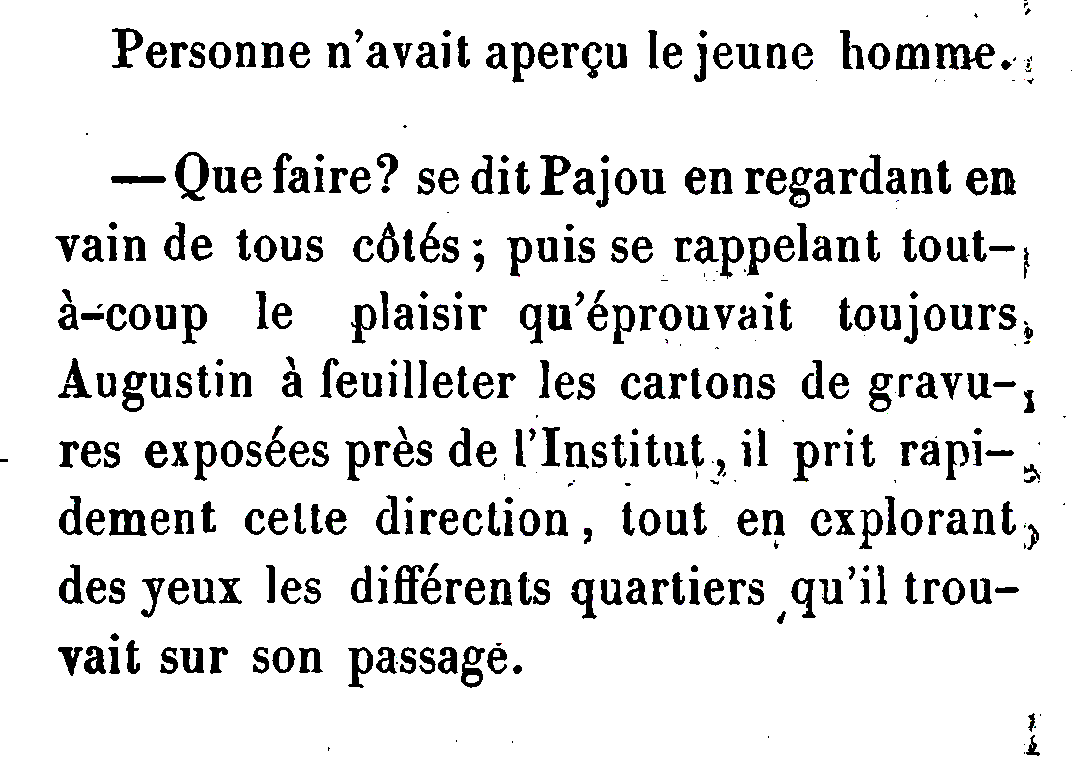}
    \\ \noalign{\smallskip}

    \includegraphics[width=0.30 \columnwidth, height=20mm]{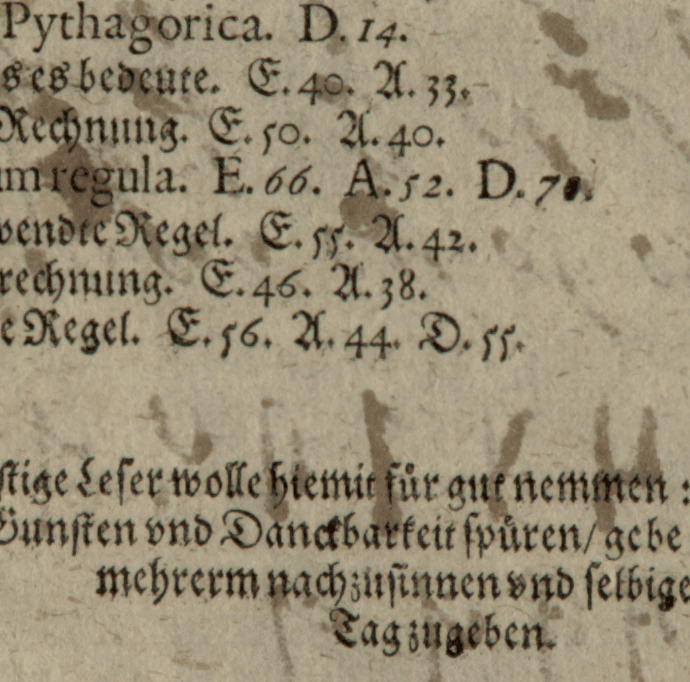} &
    \includegraphics[width=0.30 \columnwidth, height=20mm]{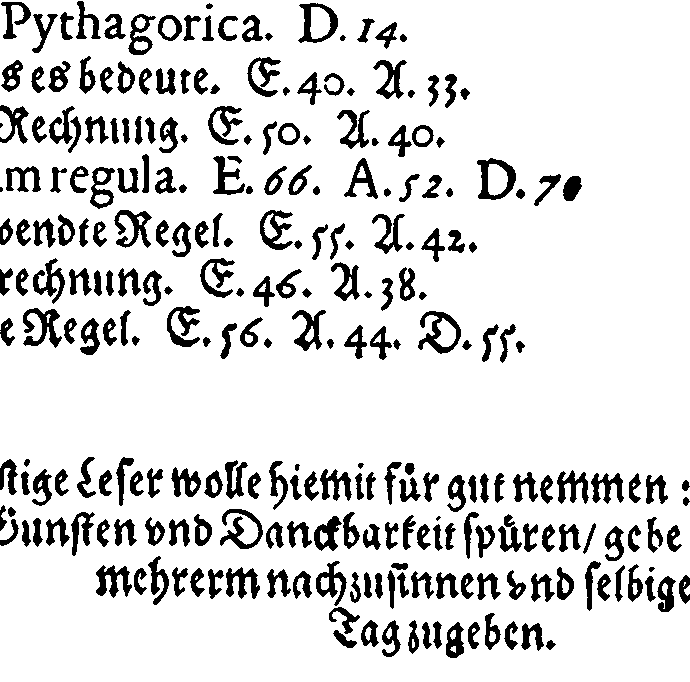}&
    \includegraphics[width=0.30 \columnwidth, height=20mm]{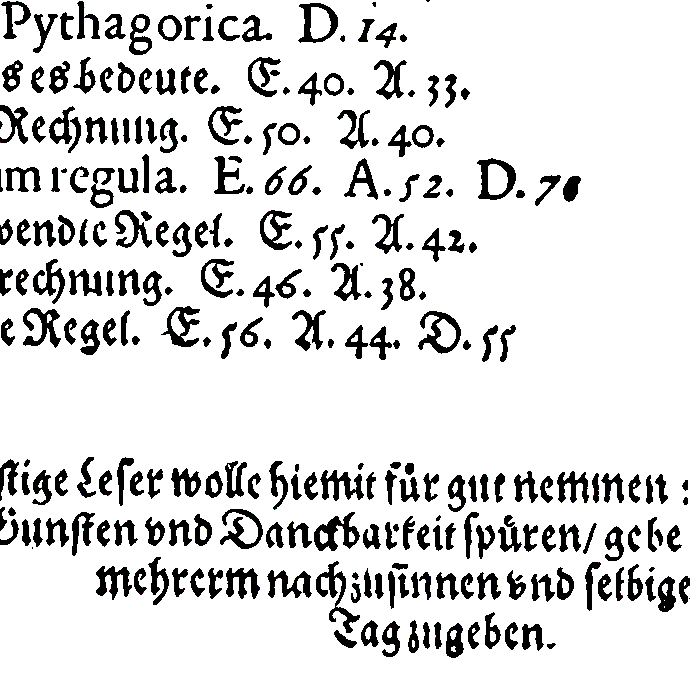}
    \\ \noalign{\smallskip}     
    
    \includegraphics[width=0.30 \columnwidth, height=20mm]{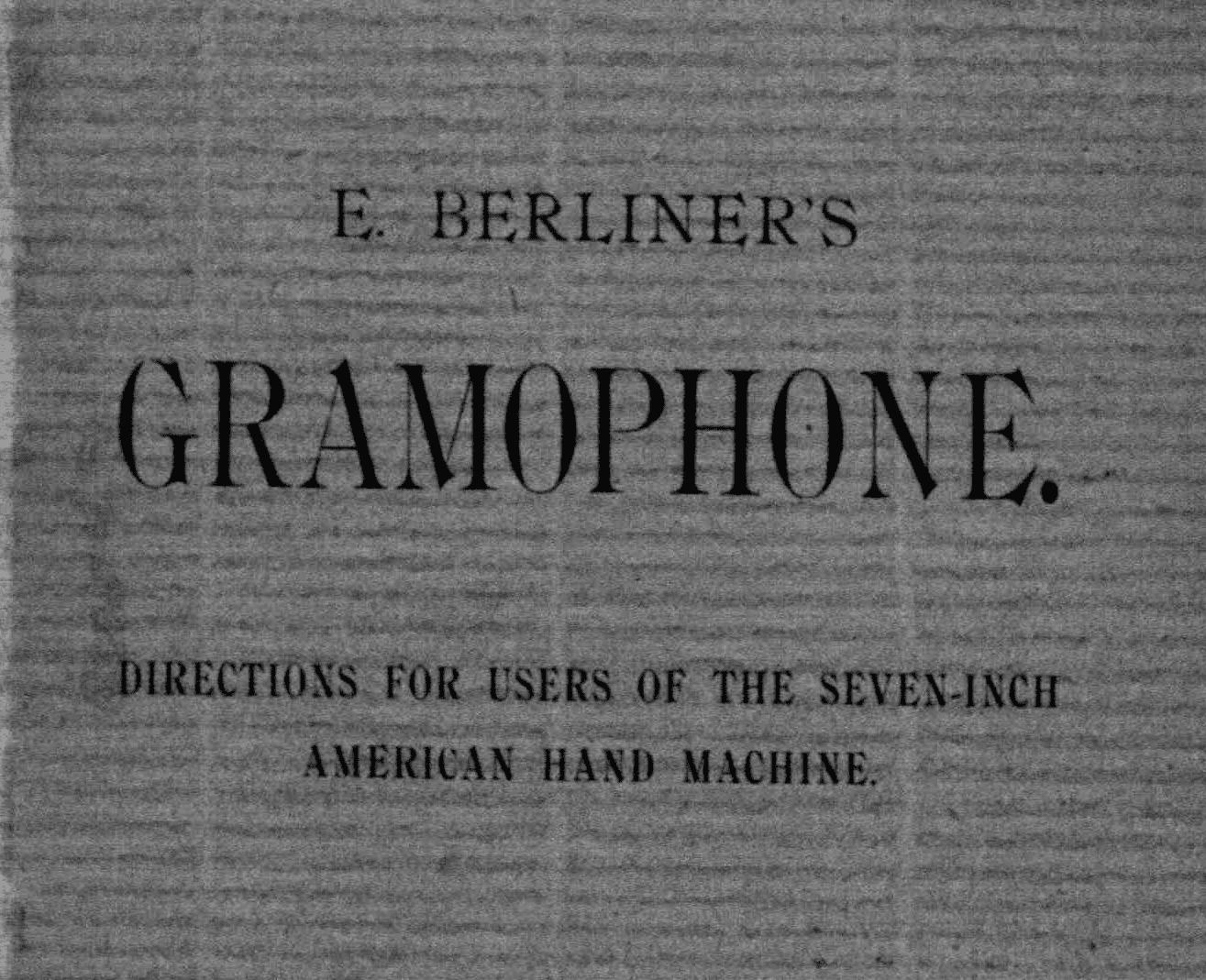} &
    \includegraphics[width=0.30 \columnwidth, height=20mm]{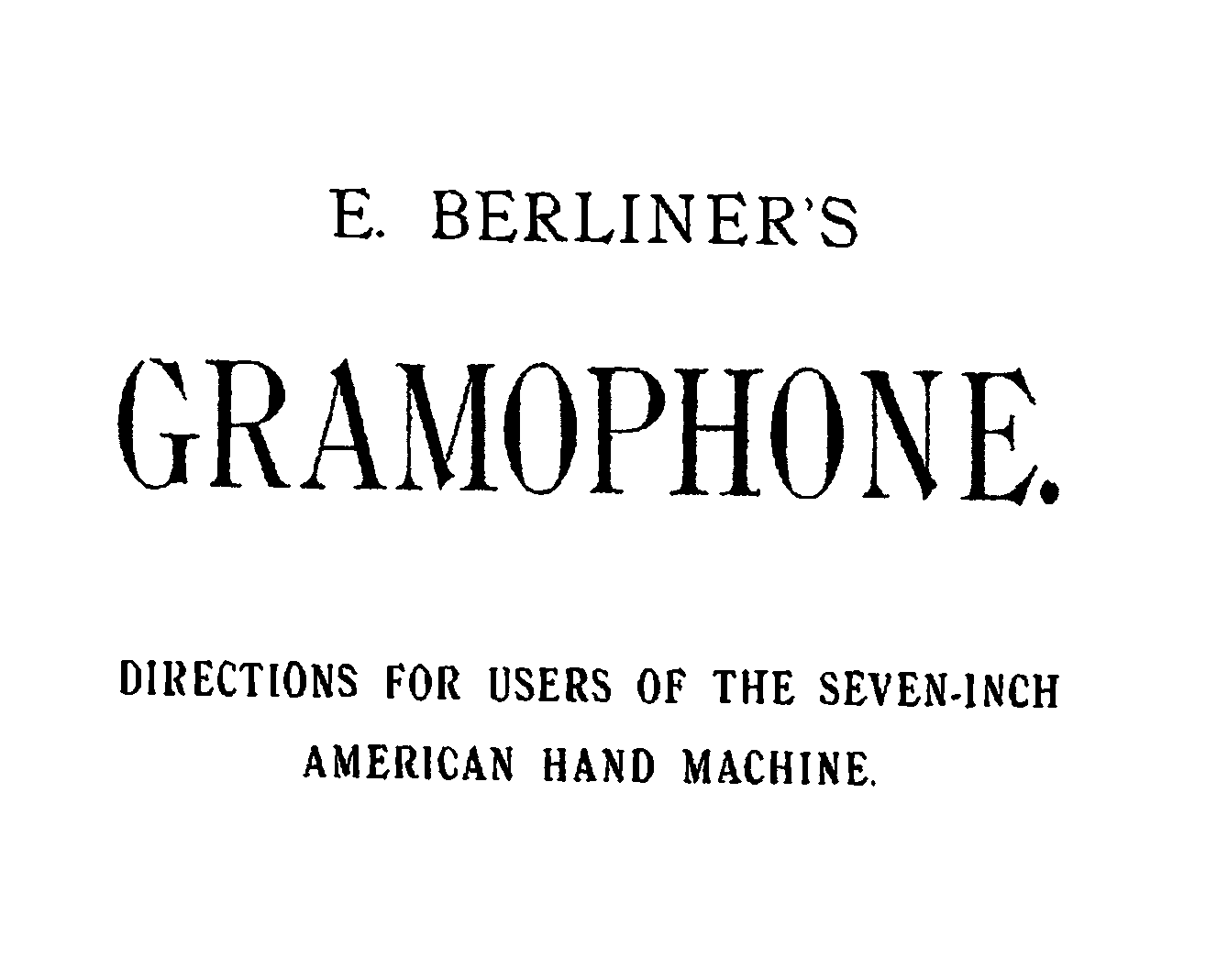}&
    \includegraphics[width=0.30 \columnwidth, height=20mm]{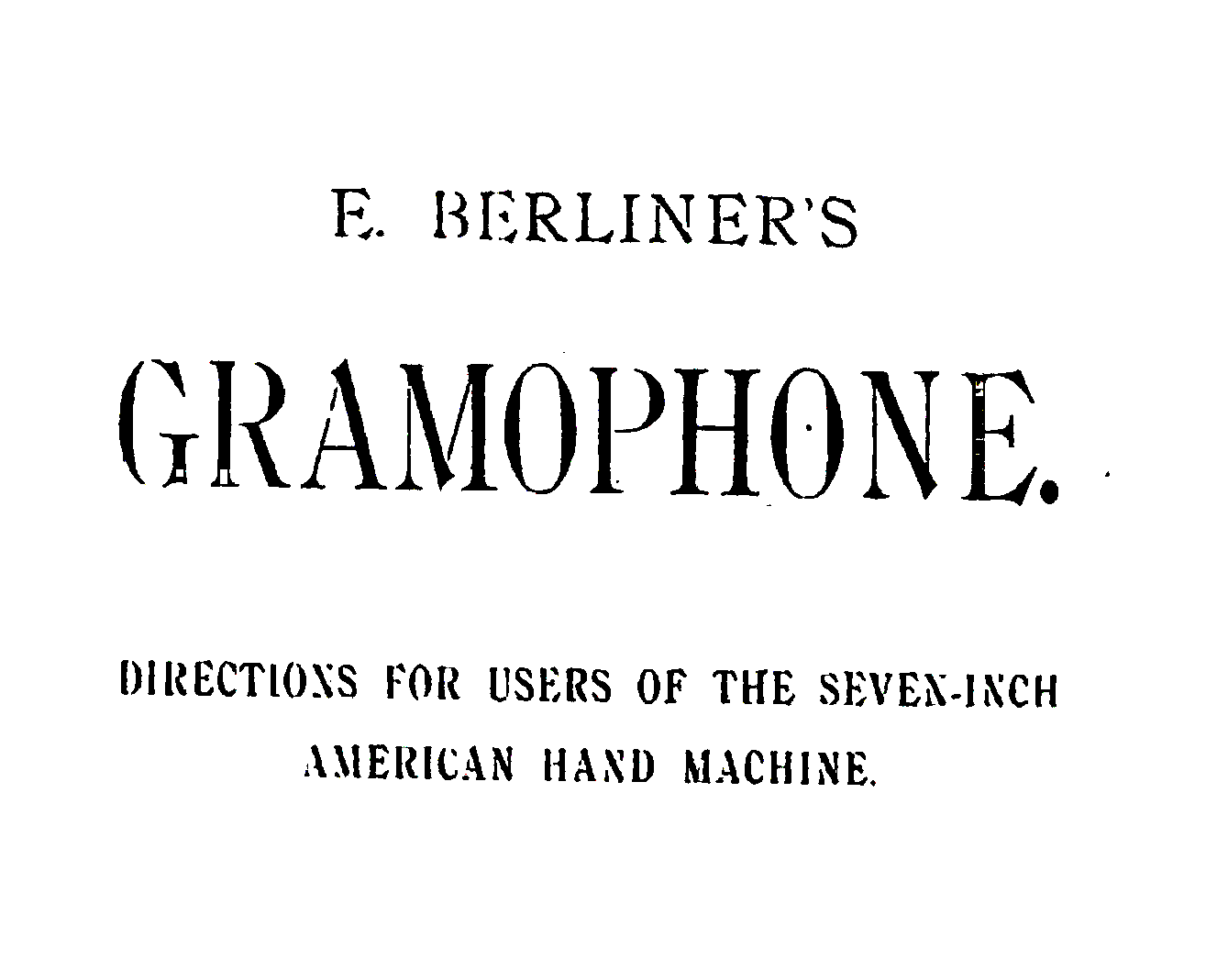}
    \\ \noalign{\smallskip}   
    
 \end{tabular}

 \caption{Qualitative results of our proposed method in binarization of some samples from the  DIBCO and H-DIBCO datasets. Images in columns are: Left: original image, Middle: GT image, Right: Binarized image using our proposed method.}
 \label{fig:our_results_dibcos_details}
\end{center}
 \end{figure}

After presenting the achieved quantitative results by our model, we present in this subsection some qualitative results. We begin by showing the enhancing performance of our method. This is illustrated in Fig.~\ref{fig:our_results_dibcos_details}, where we compare our binarization results with the GT clean images. As it can be seen, our model produces  highly clean images, which are very close to the optimal GT images, reflecting the good quantitative performance that was obtained in the previous subsection.


\begin{figure}[t]
  \begin{center}
   \begin{tabular}{cc}
 
    \includegraphics[width=0.4 \columnwidth, height=25mm]{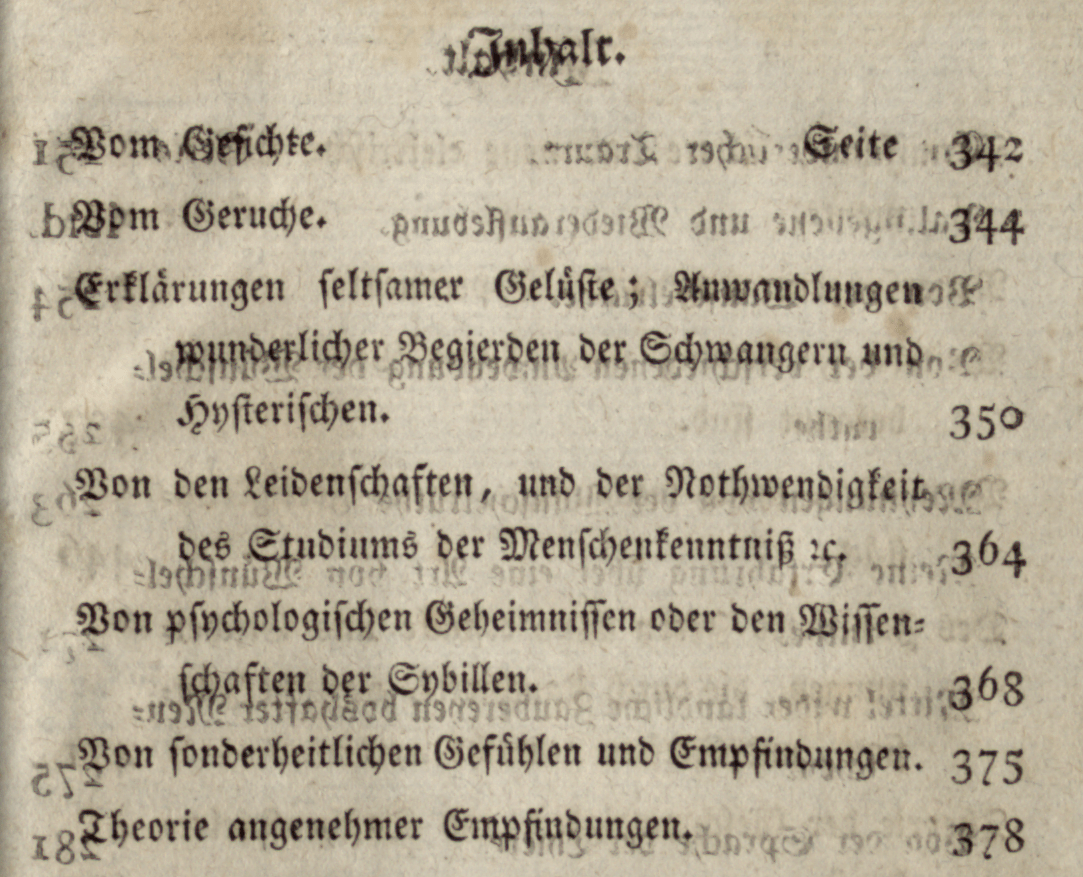} &
    \includegraphics[width=0.4 \columnwidth, height=25mm]{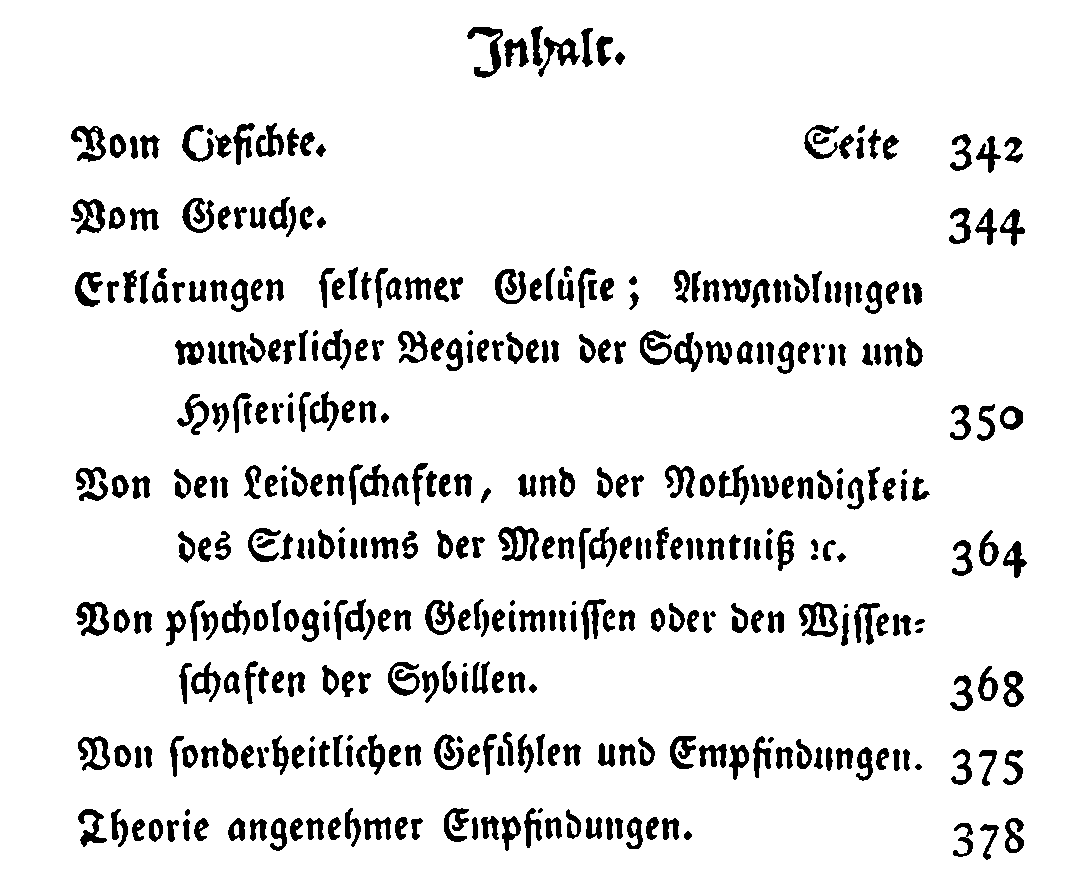}\\ \noalign{\smallskip} 
    Original  & Ground Truth 
    
    \\ \noalign{\smallskip}     
    \includegraphics[width=0.4 \columnwidth, height=25mm]{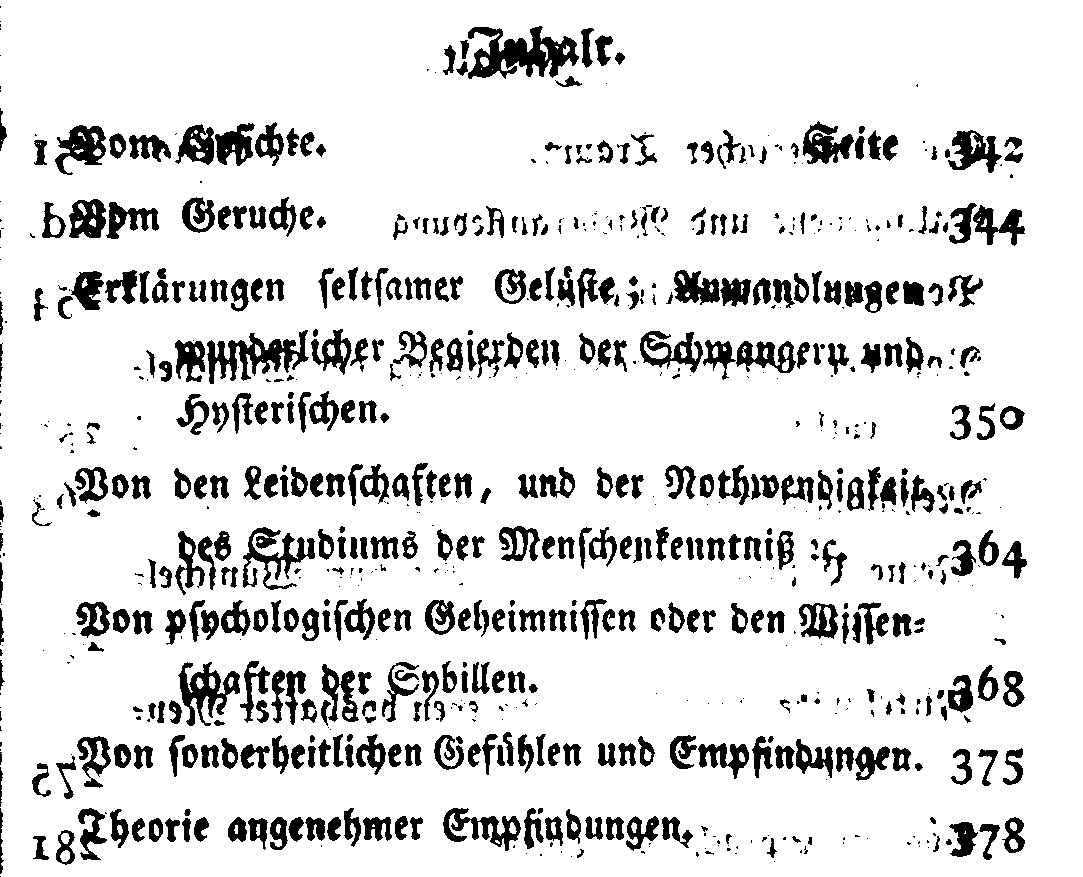}&
    \includegraphics[width=0.4 \columnwidth, height=25mm]{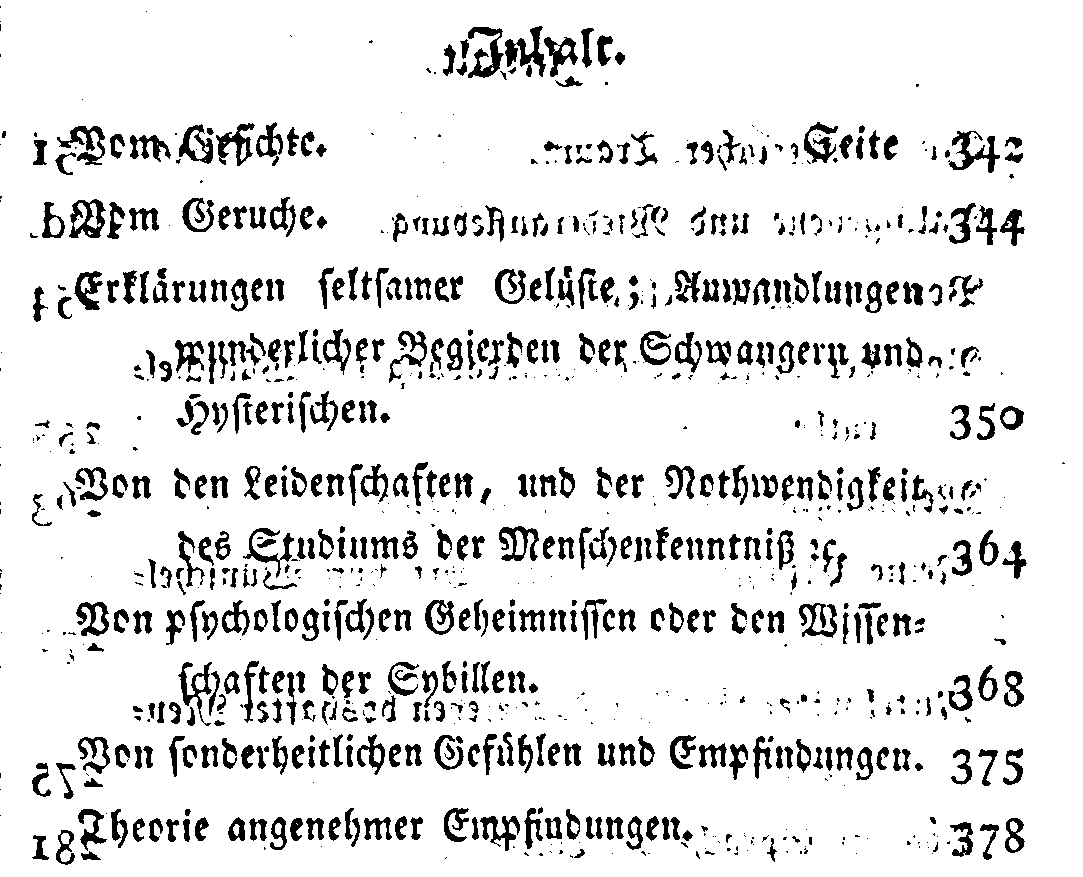} 
    \\\noalign{\smallskip}   
    Otsu \cite{otsu1979threshold} & Sauvola et al. \cite{sauvola2000adaptive}
    \\\noalign{\smallskip}

    \includegraphics[width=0.4 \columnwidth, height=25mm]{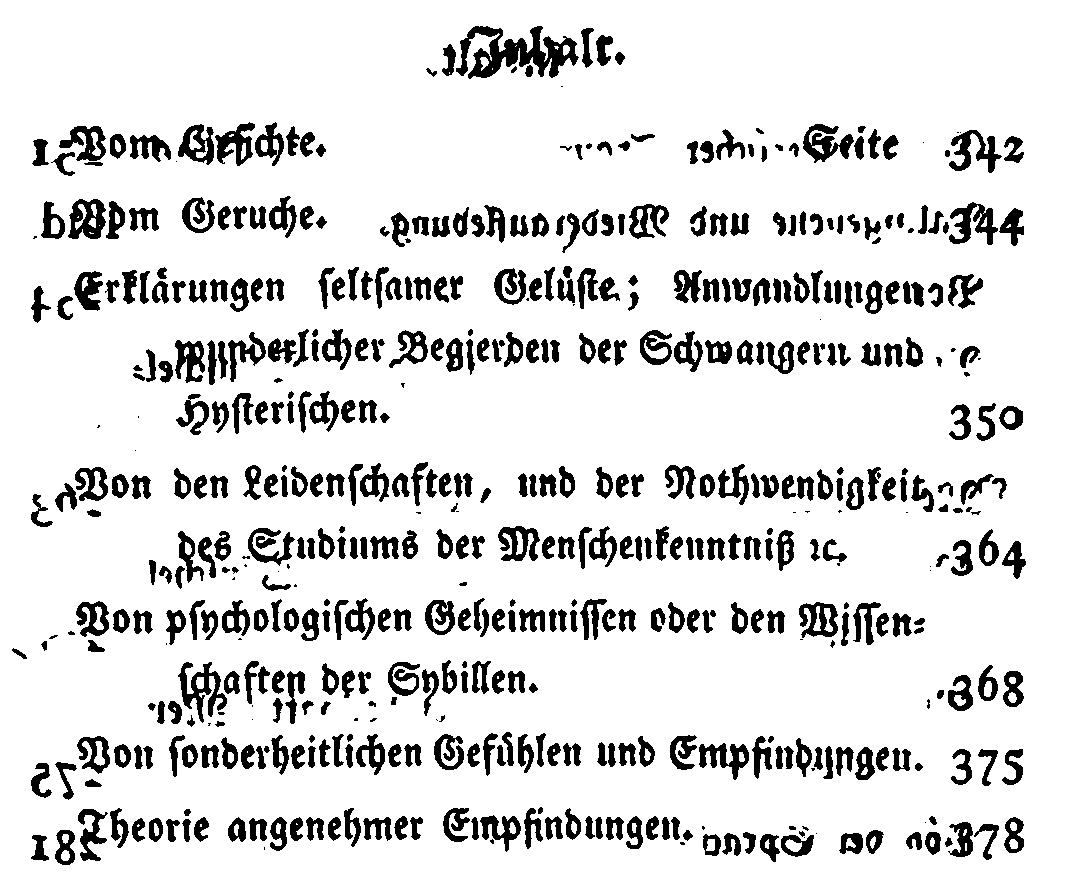}&
    \includegraphics[width=0.4 \columnwidth, height=25mm]{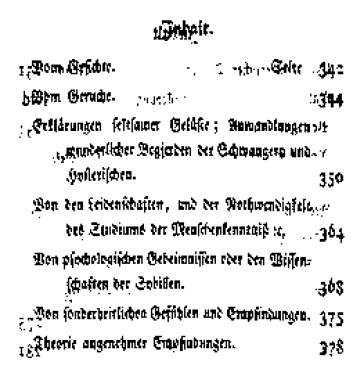} 
    \\\noalign{\smallskip}   
    Jemni et al. \cite{jemni2022enhance} & Kang et al. \cite{kang2021complex}
    \\\noalign{\smallskip}

     \includegraphics[width=0.4 \columnwidth, height=25mm]{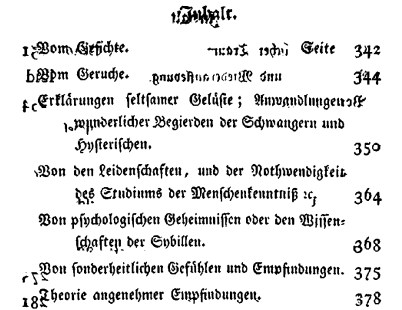}&
    \includegraphics[width=0.4 \columnwidth, height=25mm]{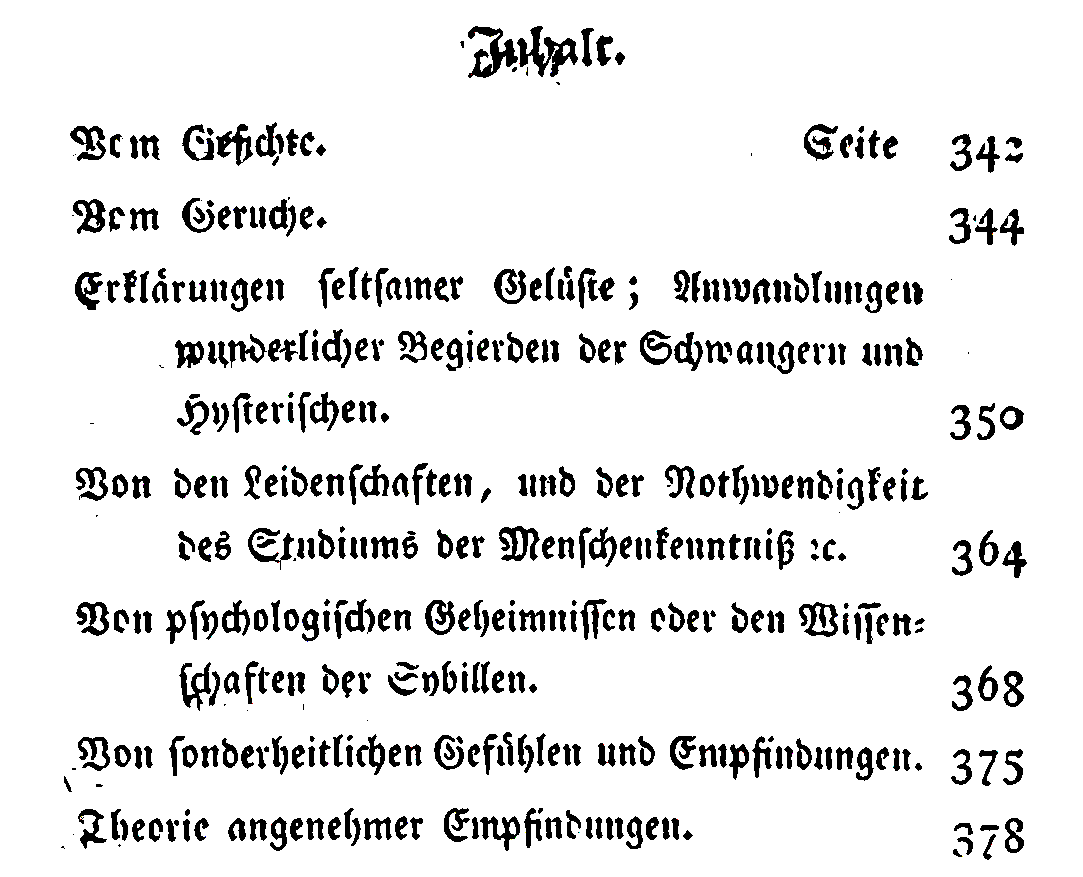} 
    \\\noalign{\smallskip}   
   Competition Winner & \textbf{Ours}
    \\\noalign{\smallskip}

 \end{tabular}

 \caption{Qualitative results of the  different binarization methods on the  sample number 12 from DIBCO 2017 Dataset.}
 \label{fig:results_dibco17}
\end{center}
 \end{figure}

 Then, we present a quantitative comparison of our method with the related approaches. This is shown in Fig.~\ref{fig:results_dibco17}, where we can notice the superiority of our model in recovering a highly degraded image over the classic thresholding \cite{otsu1979threshold,sauvola2000adaptive}, CNN \cite{kang2021complex}, and cGAN \cite{jemni2022enhance} methods.


\subsection{Self-attention Mechanism}

\begin{figure}[!t]
\centering
\begin{tabular}{|c|c|c|c|}
\hline
Degraded & Attention & Predicted & GT \\
\hline
\includegraphics[width=0.20\linewidth]{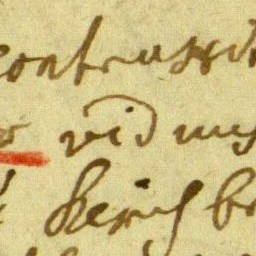} & 
\includegraphics[width=0.20\linewidth]{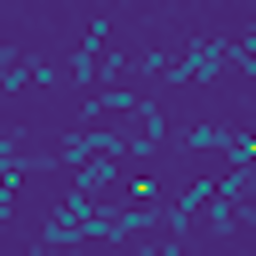}&
\includegraphics[width=0.20\linewidth]{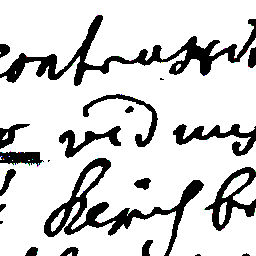} &
\includegraphics[width=0.20\linewidth]{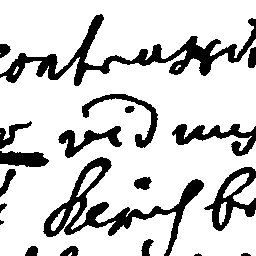}\\


\hline
\includegraphics[width=0.20\linewidth]{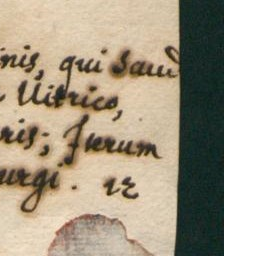} & 
\includegraphics[width=0.20\linewidth]{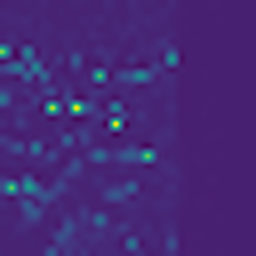}&
\includegraphics[width=0.20\linewidth]{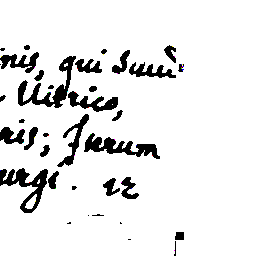} &
\includegraphics[width=0.20\linewidth]{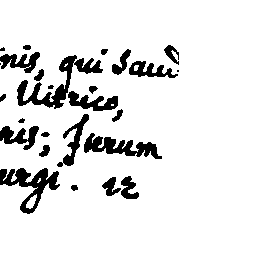}\\

\hline
\includegraphics[width=0.20\linewidth]{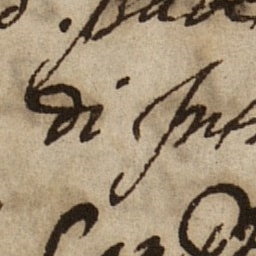} & 
\includegraphics[width=0.20\linewidth]{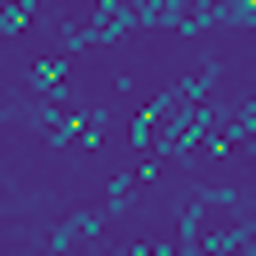}&
\includegraphics[width=0.20\linewidth]{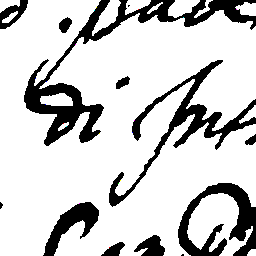} &
\includegraphics[width=0.20\linewidth]{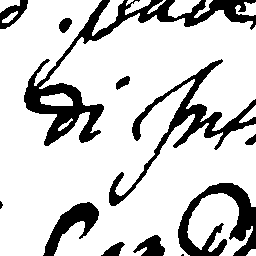}\\

\hline
\includegraphics[width=0.20\linewidth]{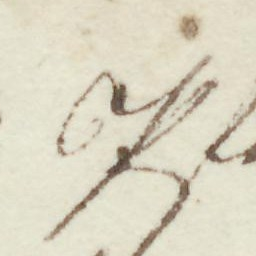} & 
\includegraphics[width=0.20\linewidth]{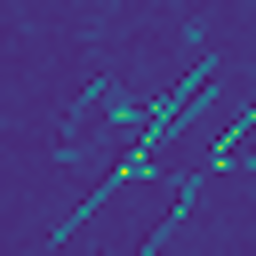}&
\includegraphics[width=0.20\linewidth]{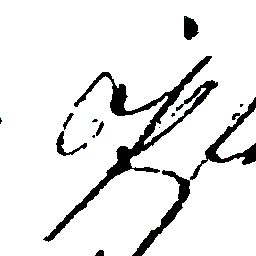} &
\includegraphics[width=0.20\linewidth]{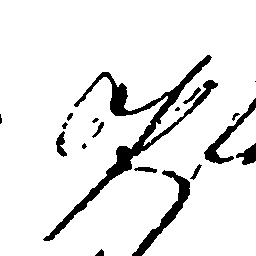}\\
\hline
\end{tabular}

\caption{Attention maps from the 2$^{nd}$ head of the last layer of  DocEnTr\{8\} encoder. We  display the self-attention for different (random) tokens.}
\label{fig:attention_success}
\end{figure}

\begin{figure}[!t]
\centering
\begin{tabular}{|c|c|c|c|}
\hline
Degraded & Attention & Predicted & GT \\
\hline
\includegraphics[width=0.20\linewidth]{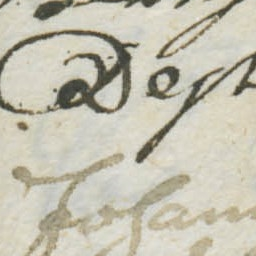} & 
\includegraphics[width=0.20\linewidth]{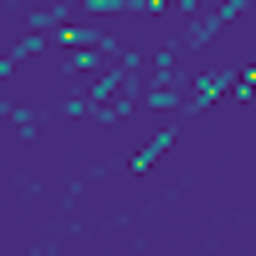}&
\includegraphics[width=0.20\linewidth]{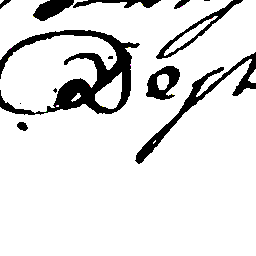} &
\includegraphics[width=0.20\linewidth]{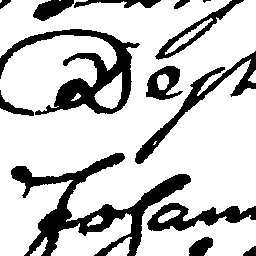}\\

\hline
\includegraphics[width=0.20\linewidth]{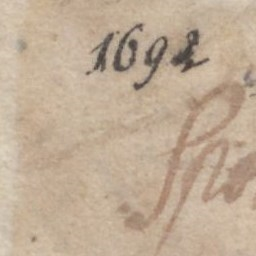} & 
\includegraphics[width=0.20\linewidth]{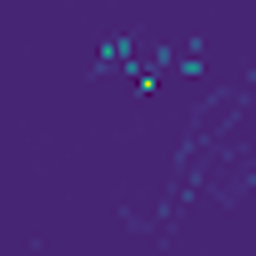}&
\includegraphics[width=0.20\linewidth]{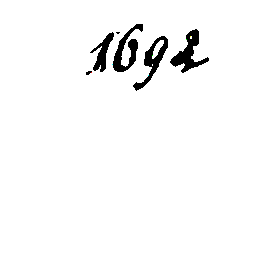} &
\includegraphics[width=0.20\linewidth]{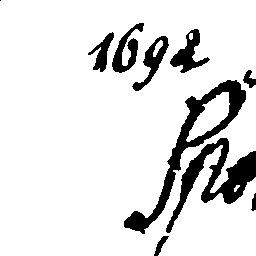}\\

\hline
\end{tabular}

\caption{Attention maps from the 2$^{nd}$ head of the last layer of  DocEnTr\{8\} encoder. We  display the self-attention for different (random) tokens. (A failure case).}
\label{fig:attention_failure}
\end{figure}

As we stated above, our method differs from the CNN related ones by employing the transformers to enhance the degraded document images. The self-attention mechanism used in the transformer blocks gives a global view to every token on the other tokens that represents the  patches within the image for a better enhancing result. A visual illustration of the attention maps of the last layer from the encoder is given in Fig.~\ref{fig:attention_success}. As it can be seen, a token can attend to all the patches within the image. In these test cases each token (patch representation) is focusing on the text elements, while ignoring the degraded patches.  Thus, the attending patches are decoded later and projected to pixels while taking into consideration a high-level global information from the attended neighbouring patches that cover the full input image.  We also notice that the attention maps are mostly matching the text of the GT images, which lead to a satisfactory binarization result that is closer to the GT. This supports the utility of using the transformers with its powerful self-attention mechanism in the image enhancement task. However, in other sample cases as illustrated in Fig.~\ref{fig:attention_failure},  we observe that the attention maps are considering some portions of the text as a background region. Hence, the resultant enhanced image is removing foreground text because it considers it as a background noise. This explains the failure of the self-attention paradigm in these scenarios.

\section{Conclusion}\label{s:conclusion}
This paper presents a novel transformer-based architecture called DocEnTr for document image enhancement. To the best of our knowledge, this is the first pure transformer model addressing DIE related problems. 
The model captures high-level global long-range dependencies using the self-attention mechanism for a better performance.  Quantitative and qualitative results on the DIBCO benchmarks prove the effectiveness of  DocEnTr in recovering highly degraded document images. It is a simple and flexible framework that can also be easily applied to enhance other kinds of degradation occurring in document images (like blur, shadow, warps, stains etc). These aspects will be investigated in a future work. We also wish to investigate a self-supervised learning stage that can substantially benefit from large amounts of unlabeled data.

\section*{Acknowledgment}

This work has been partially supported by  the Swedish Research Council (grant 2018-06074, DECRYPT), the Spanish projects RTI2018-095645-B-C21, the CERCA Program / Generalitat de Catalunya, the FCT-19-15244,  the Catalan projects 2017-SGR-1783, PhD Scholarship from AGAUR (2021FIB-10010) and DocPRESERV project (Swedish STINT grant).

{\small
\bibliographystyle{ieeetran}
\bibliography{main.bib}
}

\end{document}